\documentclass[10pt,twocolumn,letterpaper]{article}

\usepackage{cvpr}
\usepackage{times}
\usepackage{epsfig}
\usepackage{graphicx}
\usepackage{amsmath}
\usepackage{amssymb}

\graphicspath{{./figs/}}

\usepackage{tabularx}

\newcommand{\norm}[1]{\left\| {#1} \right\|}

\newcommand{\R}{\ensuremath \mathbb{R}}

\renewcommand{\epsilon}{\varepsilon}
\def\utilde#1{\mathord{\vtop{\ialign{##\crcr
$\hfil\displaystyle{#1}\hfil$\crcr\noalign{\kern1.5pt\nointerlineskip}
$\hfil\tilde{}\hfil$\crcr\noalign{\kern1.5pt}}}}}
\newcommand{\hide}[1]{}

\usepackage[pagebackref=true,breaklinks=true,letterpaper=true,colorlinks,bookmarks=false]{hyperref}

\cvprfinalcopy % *** Uncomment this line for the final submission

\ifcvprfinal\fi
\begin{document}

%%%%%%%%% TITLE

\title{Hierarchical Recurrent Attention Networks for  Structured Online Maps }

\author{
	Namdar Homayounfar$^{1,2}$ \quad Wei-Chiu Ma$^{1,3}$, Shrinidhi Kowshika Lakshmikanth$^{1}$,\quad Raquel Urtasun$^{1,2}$\\
	$^{1}$Uber Advanced Technologies Group \quad $^{2}$University of Toronto \quad $^{3}$ MIT\\
	%	\small\texttt{\{namdar,weichiu,justin.liang,xinyuw,jackwindows,urtasun\}@uber.com}
	\small\texttt{namdar.homayounfar@mail.utoronto.ca,  weichium@mit.edu, kowshika@cs.cmu.edu} \\ 
	\small\texttt{urtasun@cs.toronto.edu}
}

\maketitle
%\thispagestyle{empty}

%%%%%%%%% ABSTRACT
\begin{abstract}
	In this paper, we tackle the problem of online road network extraction from sparse 3D point clouds. Our method is inspired by how an annotator builds a lane graph, by first identifying how many lanes there are and then drawing each one in turn. 
	We develop a hierarchical recurrent network that attends to initial regions of a lane boundary and traces them out completely by outputting a structured polyline. We also propose a novel differentiable loss function that measures the deviation of the edges of the ground truth polylines and their predictions. This is more suitable than distances on vertices, as  there exists many ways to draw equivalent polylines. We demonstrate the effectiveness of our method on a 90 km stretch of highway, and show that we can recover the right topology 92\% of the time.  
\end{abstract}

%%%%%%%%% Introduction
% !TEX root = top.tex

\section{Introduction}

A self driving car software stack is typically composed of three main components: perception, prediction and motion planning.{ \it Perception} consists of estimating where everyone is in the scene in 3D given data from the sensors (e.g., LIDAR, cameras, radar, ultrasonic). {\it Prediction} is concerned with predicting the future action of the dynamic objects that have been identified by perception. Finally, the outputs of perception and prediction are used by {\it motion planning} in order to decide the trajectory of the ego-car. 

Despite several decades of research, these three tasks remain an open problem. To facilitate these tasks, most self-driving car teams rely on  
high definition  maps, commonly referred to as {\it HD maps}, which contain both geometric and semantic information about the static environment. 
For example, planning where to go is easier if we know the geometry of the road (i.e., lane composition). This information is also very useful to determine the future motion of other traffic participants. 
Furthermore,  false positives in vehicle detection can be reduced if we know where the road is.

To create these maps, most self-driving car programs rely on offline processes where semantic components such as  lanes are extracted with the help of a user in the loop. Furthermore, they typically require multiple passes over the same environment in order to build accurate geometric representations. This  is very expensive and requires a dedicated fleet for mapping. 
It has been estimated that mapping the US only once will cost over 2 Billion dollars. This approach is not scalable globally and thus it is of fundamental importance to design online mapping algorithms that do not require a user in the loop or in the least minimize their involvement to correction tasks.

\begin{figure}[t]
\begin{center}
   \includegraphics[width=\linewidth]{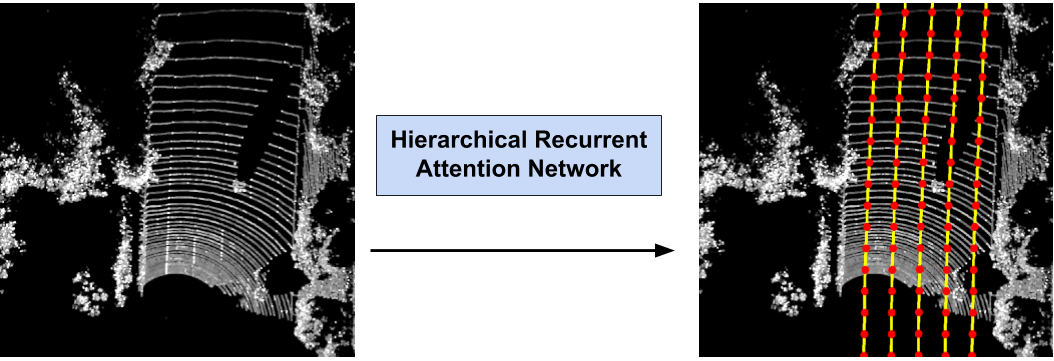}
\end{center}
   \caption{Our hierarchical recurrent attention model takes a sparse point cloud sweep of the road (\textbf{right}) and outputs (\textbf{left}) a structured representation of the road network where each lane boundary instance is retrieved.}
   \label{fig:intro}
\end{figure}

The most basic information that is required for driving is to be able to extract the location of the lanes in 3D space (mapping), and their relationship to the ego-car (localization to the map). 
In the context of maps, these  lanes are structured objects  and  are typically represented as a set of polylines, one per lane boundary.  We refer the reader to Fig. \ref{fig:intro} for a representation of a lane graph.

Existing automatic approaches to lane extraction have two major drawbacks. 
First, they treat the problem as semantic segmentation \cite{mattyus2015enhancing, mattyus2016hd} or lane marking detection \cite{lee2017vpgnet,huval2015empirical, kim2014robust}. As a consequence they produce solutions that are not topologically correct, e.g., a lane might have holes due to occlusion. This is problematic as most motion planners can only handle lane graphs that are structured and represent the right topology. 
Second, these methods attempt to extract lane graphs from camera images \cite{yoo2017robust}.  Unfortunately, a good image based lane estimation is not equivalent to an accurate 3D lane extraction. Due to perspective projection, pixels in image space have different physical widths in 3D. This largely limits their applications in real world.

With these challenges in mind, we present a novel approach to online mapping that extracts structured lane boundaries directly from a single LIDAR sweep. 
To be more specific, we propose a hierarchical recurrent neural network that is able to both count the number of lanes and draw them out.  
The model takes as input sparse LIDAR point clouds, which is the natural 3D space for detecting lane boundaries, and outputs a structured representation of the road network that is topologically correct and hence consumable by existing motion planners. 
As there exists many ways to draw equivalent polylines, we further develop a novel differentiable loss function that directly minimizes the distance between two polylines rather than penalizing the deviation of GT and predicted vertices using cross-entropy or regression loss \cite{CastrejonCVPR17}. The objective helps our network focus on learning the important lane graph structure, rather than the irrelevant vertices coordinates. Our model can be trained in an end-to-end manner without heuristics or post-processing steps such as curve fitting.
The overall process mimics how humans annotate maps and thus is amenable to a guided user in the loop.

We demonstrate the effectiveness of our approach on highway sequences captured over a range of 90 km. Our approach determines the right topology 92\% of the time. Furthermore, we recover the correct road network with an average of 92\% precision and 91\% recall at a maximum distance of 20 cm away from the lane boundaries .

% !TEX root = top.tex

\section{Related Work}

\paragraph{Road and Lane Detection}

Finding the drivable path in front of an autonomous vehicle and the lane boundaries is of outmost importance. \cite{mohan2014deep, levi2015stixelnet, yao2015estimating} apply graphical models on manually generated annotations to estimate the free space and the road. Some other methods \cite{tan2006color, lieb2005adaptive, paz2015variational, alvarez2011road,kong2010general,cheng2006lane,wedel2009b, alvarez2012road, kuhnl2012spatial} use unsupervised or self-supervised methods based on color, texture or geometric priors to detect the road. The authors in \cite{laddha2016map, wang2015holistic,irie2013road,alvarez2014combining,mnih2012learning}  develop road detection algorithms either by automatically or manully generating labels from maps or using them as priors. The authors in \cite{fritsch2013new} extend a small subset of the KITTI \cite{kitti} dataset and provide a benchmark for detecting polygons that define the free space of the road and the ego-lane. Recently, the resurgence of deep learning methods \cite{alexnet, shmidoverview} has provided tremendous success  in many different computer vision tasks. For lane detection, \cite{lee2017vpgnet} train a neural network that detects land and road markings by leveraging vanishing points.

\paragraph{Semantic segmentation of aerial images/ road network extraction}
Aerial imagery can be used for road network extraction and segmentation. 
Although, aerial imagery can cover a huge portion of the world, they operate on a lower resolution and thus cannot be used for fine grained map creation. \cite{mattyus2015enhancing, mattyus2016hd} enhance freely available maps using  aerial imagery by fine grained semantic segmentation and inference in an MRF. 
In other work \cite{wegner2013higher, wegner2015road, montoya2014mind} extract the road network from aerial images using a CRF model. \cite{marmanis2016semantic} use an end-to-end fully convolutional network to segment high resolution aerial images. \cite{marmanis2016semantic} presents an end-to-end semantic segmentation deep learning approach of very high resolution aerial images. Recently, The Torontocity dataset \cite{TCity2017} provides a benchmark for extracting road curbs and centerlines from bird's eye view maps.

\paragraph{Other} 
Our work is inspired by \cite{CastrejonCVPR17} in which the authors develop a semi-automatic annotation tool of object instances by directly predicting the vertices of the polygon outlining the object's segmentation mask. They use the cross-entropy loss to learn the position of the vertices. We note that a loss function on the location of the vertices is not appropriate as there are many ways to draw the same polygon. As such, we design a novel loss function that penalizes directly in a differentiable manner the deviation of the edges of the predicted polylines from their ground truth. 
In \cite{ren17recattend,romera2016recurrent} the a recurrent network iteratively attends to object instances and segments them. We use a similar idea by attending to lane boundaries and drawing them out.

% !TEX root = top.tex

\section{Hierarchical Recurrent Attention Model for Lane Graph Extraction}

\begin{figure}[t]
\begin{center}
   \includegraphics[width=\linewidth]{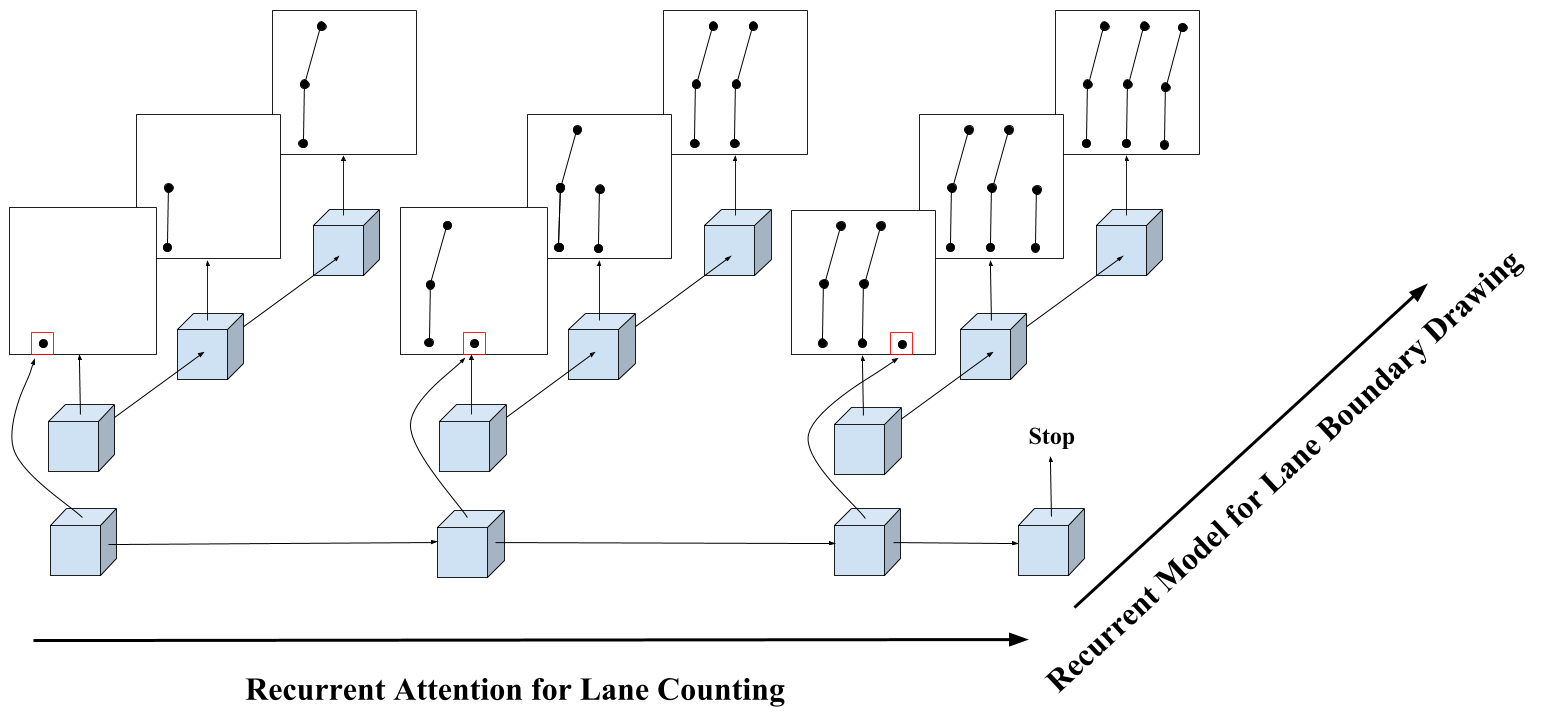}
\end{center}
\caption{The overall structure of our model where a convolutional recurrent neural network sequentially attends to the initial regions of the lane boundaries while a another convolutional LSTM traces them out fully. This process iterates until the first RNN signals a stop. }
   \label{fig:model_over}
\end{figure}

Our goal  is to extract a structured representation of the road network, which we will henceforth refer to as a {\it lane graph}. A lane graph is defined as a set of polylines, each of which defines a lane boundary.  
Towards this goal, we exploit a LIDAR sweep in the form of a point cloud projected onto bird's eye view (BEV) as our input $x \in \R^{H\times W \times 1}$. The point cloud contains LIDAR intensity for each point, a cue that allows us to exploit the reflective properties of paint. 
This provides us with a sparse representation of the 3D world. 
We refer the reader to Fig. \ref{fig:intro} for an example of our point cloud input and the  predicted lane graph.

Our approach is inspired by how humans create lane graphs when building maps. In particular, annotators are presented with a bird's eye view of the world,  and sequentially draw one lane boundary at a time. To do so, they typically start from the bottom corner of the left most lane and draw each lane by first choosing  an initial vertex on the lane boundary and tracing the lane boundary  by a sequence of further clicks. 

When the lane boundary is fully specified in the form of a polyline, the annotator moves on to its closest neighbour and repeats the process until no more lane boundaries are visible.       

We design a structured neural network that closely mimics this process as demonstrated in Fig. \ref{fig:model_over}. Our hierarchical recurrent network sequentially produces a distribution over the initial regions of the lane boundaries, attends to them and then draws a polyline over the chosen lane boundary by outputting a sequence of vertices. 

Our network  iterates this process until it decides that no more lane boundaries are present and it is time to stop.

 In the following, we explain in detail the main components of our model.
In particular, an encoder network is shared by a recurrent attention module that attends to the initial regions of the lane boundaries (Section \ref{sec:rnn_count}) and a decoder network that feeds into a conv-lstm that draws each lane boundary given the initial region (Section \ref{sec:rnn_poly}).

\subsection{Encoder-Decoder Backbone} \label{sec:backbone}
Our model is based upon the feature pyramid networks of \cite{lin2016feature,lin2017focal,linknet},  where a residual \cite{resnet} encoder-decoder architecture with lateral additive connections is used to build features at different scales. 
The features of the encoder are shared by both the recurrent attention module \ref{sec:rnn_count} and the Polyline-RNN \ref{sec:rnn_poly} and capture information about the location of the lane boundaries at different scales. 
The decoder is composed of multiple convolution and bilinear upsampling modules that build a feature map used by only the Polyline-RNN module. We use batch norm \cite{ioffe2015batch} and ReLU non-linearity throughout the network. The exact architecture is outlined in the supplementary material.

\begin{figure*}[t]
\begin{center}
   \includegraphics[width=\linewidth]{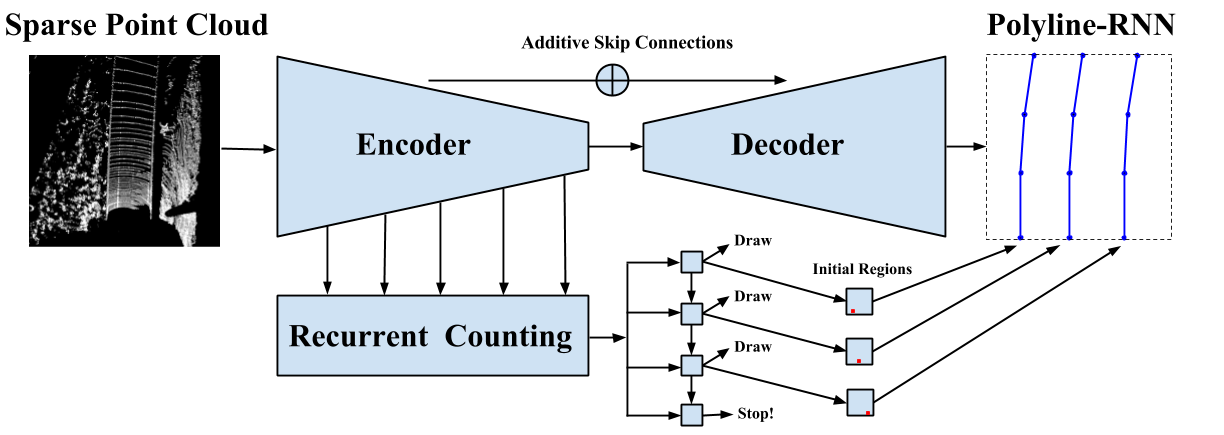}
\end{center}
   \caption{Our Hierarchical Recurrent Attention Network: An encoder network is shared by  the recurrent attention module for counting and attending to the initial regions of the lane boundaries as well as a decoder that provides features for the Polyline-RNN module that draws the lane boundaries of the sparse point cloud.}
   \label{fig:model}
   \vspace{-5mm}
\end{figure*}

\subsection{Recurrent Attention  for Lane Counting} \label{sec:rnn_count}

\begin{figure}[!t]
\[\arraycolsep=1.0pt
\begin{array}{ll}
\includegraphics[width=0.49\linewidth]{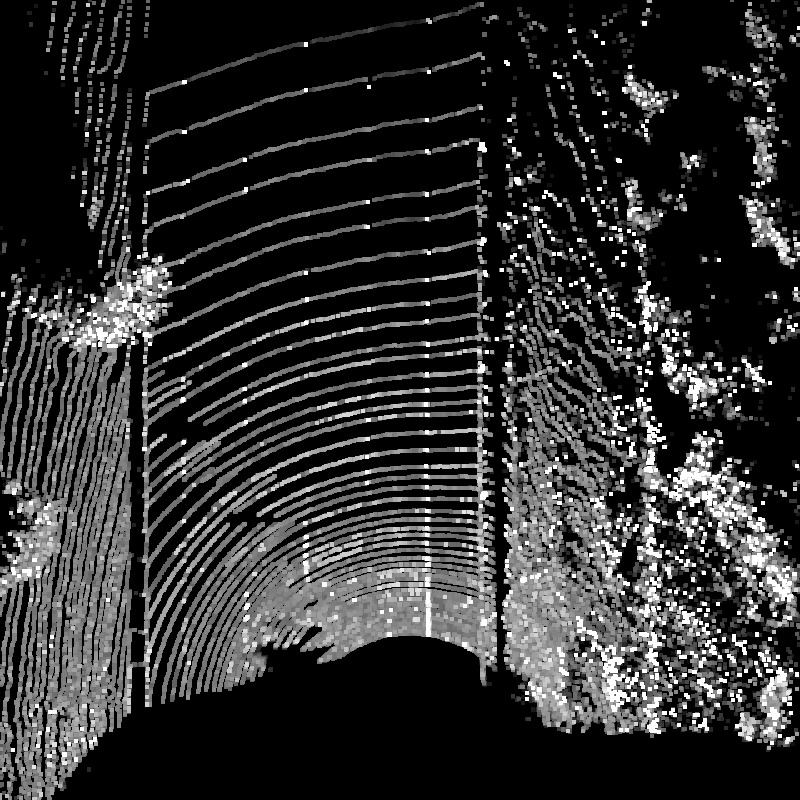}& 
\includegraphics[width=0.49\linewidth]{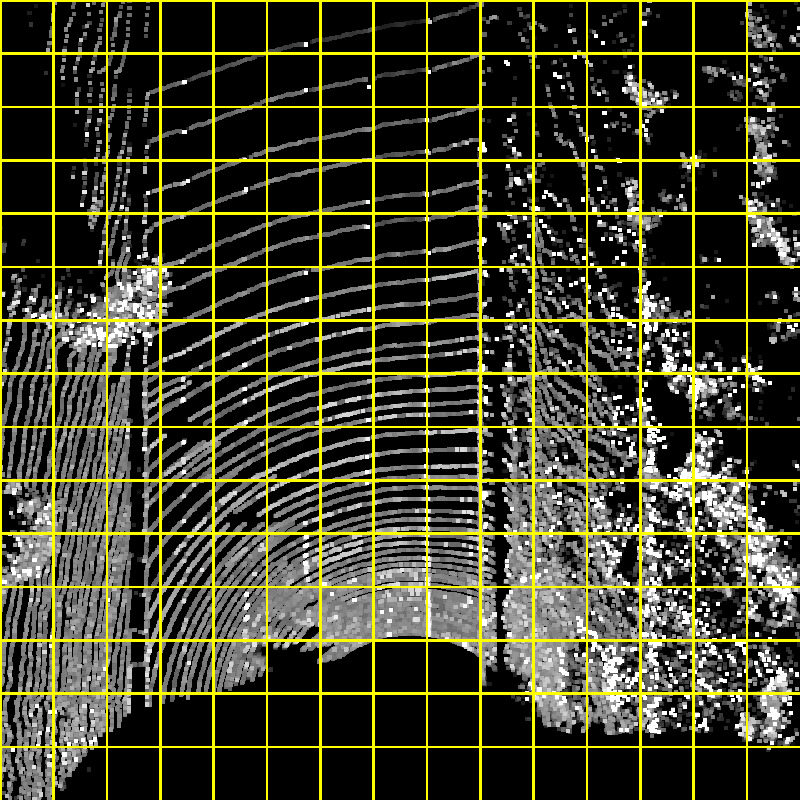}
\end{array}
\]
\caption{The input point cloud (\textbf{right}) is discretized into $K$ bins along each dimension (\textbf{left}). The recurrent attention module for counting, as well as an annotator, need only to focus on an initial region of the lane boundary rather than an exact vertex.}
\label{fig:grid2}
\end{figure}

We design a network that is able to decide how many lane boundaries exist and attends to the region in the image where the lane boundary begins. We have deliberately simplified the output of the net to be a region rather than the exact coordinates of a vertex. This way, if run in interactive mode, an annotator is only required to provide a coarse starting region for a lane boundary to be drawn. The task of predicting the actual vertex coordinates falls upon the Polyline-RNN which we shall describe in detail in section \ref{sec:rnn_poly}. These regions correspond to non-overlapping bins that are obtained by dividing the input $x$ into $K$ segments along each spatial dimension as demonstrated in Fig. \ref{fig:grid2}.

For the network to predict the starting regions of the lane boundaries, we deploy a similar strategy as \cite{pinheiro2016learning,CastrejonCVPR17} and concatenate the feature maps of the encoder network so that the net has clues at different granularities. We use convolution layers with large non-overlapping receptive fields to downsample the larger feature maps and use bilinear upsampling for the smaller feature maps to bring all of them to the same resolution. Next, this feature map is fed to two residual blocks in order to obtain a final feature map $f$ of smaller resolution than the point cloud input $x$ to the network. We reduce the resolution since we care only about the regions where a lane boundary starts rather than its exact starting coordinate.  

Next, a vanilla convolutional RNN is iteratively applied to $f$ with the task of attending to regions of $x$ and outputting a starting region of the lane boundary. 

In order to be able to stop, this RNN also outputs a binary variable denoting whether we have already counted all the lanes or not. 
In particular, at each timestep $t$, the conv-RNN outputs on one hand the probability $h_t$ of halting while the other output is a softmax $s_t$ of dimension ${K \times K \times 1}$ over the region of the starting vertex of the next lane boundary. At inference time, we replace the softmax with an argmax and threshold the probability of halting.

\subsection{Drawing Lane Boundaries with Polyline-RNN} \label{sec:rnn_poly}

We use a convolutional LSTM to  iteratively draw a polyline as a sequence of vertices. 
In particular, the recurrent attention mechanism of the previous section provides us with a region which contains the first vertex of the lane boundary. A section of dimension $H_c \times W_c$ around this region is cropped from the output feature map of the decoder of section \ref{sec:backbone} and fed into the conv-LSTM. The conv-LSTM then produces a softmax over the position of the next vertex $p_1^t$ on the lane boundary. 
The vertex $p_1^t$ is then used to crop out the next region and the process continues until the lane boundary is fully traced or we reach the boundary of the image. 
As such we obtain our polyline $P_t$ at timestep $t$.

%!TEX root = ./top.tex

\section{Learning}

To facilitate learning, we derive a multitask objective that provides supervision for each component of our hierarchical recurrent attention model. 
Our output loss function computes the difference between two polylines, the ground truth and our prediction. Note that a loss function on the location of the vertices is not appropriate as there are many ways to draw the same polyline which will have very different location of vertices. 

Instead, we directly  penalize the deviations of the two curves. 
We further provide supervision at the level of our attention mechanism over regions which contain a starting vertex of a polyline. We also define a loss function that teaches the network when to stop counting the polylines.

\subsection{Polyline Loss}

We encourage  the edges of a prediction $P$ to superimpose perfectly on those of a ground truth $Q$. In particular, we define:
\begin{align}
	L(P, Q) &=  L_{P\to Q} + L_{Q\to P} \notag\\
	&= \sum_{i}\min_{q \in Q}{ \norm{p_i - q}_2} + \sum_{j} \min_{p \in P}{ \norm{p - q_j}_2} \label{eq:poly_loss}
\end{align}
Note that although our predictions are vertices of $P$, in the above equation we sum over the coordinates of all the edge pixels of $P$ and $Q$ rather than solely their vertices. 

We note that the two terms of the loss function are symmetric. Intuitively, the first term $L_{P\to Q}$ encourages the predicted polyline $P$ to lie on the ground truth $Q$ by summing and penalizing the deviation of all the edge pixels of $P$ from those of $Q$. While necessary, this loss is not sufficient for $P$ to cover $Q$ completely since it ignores those superimposing polylines $P$ that are shorter than $Q$. 
To overcome this, the second loss $L_{Q\to P}$ instead penalizes the deviations of the ground truth from the predicted polyline. In particular, if a segment of $Q$ is not covered by $P$, all the edge pixels of that segment would incur a loss.

As noted, the loss function in Eq. \eqref{eq:poly_loss} is defined w.r.t. to all the edge pixel coordinates on $P$ whereas the Polyline-RNN network predicts only a set of vertices. As such, for every two consecutive vertices $p_j$ and $p_{j+1}$ on $P$, we obtain the coordinates of all the edge pixel points lying in-between by taking their convex combination.
This makes the gradient flow from the loss functions to the network through every edge point.

\begin{figure}[tb]
\begin{center}
   \includegraphics[width=0.9\linewidth]{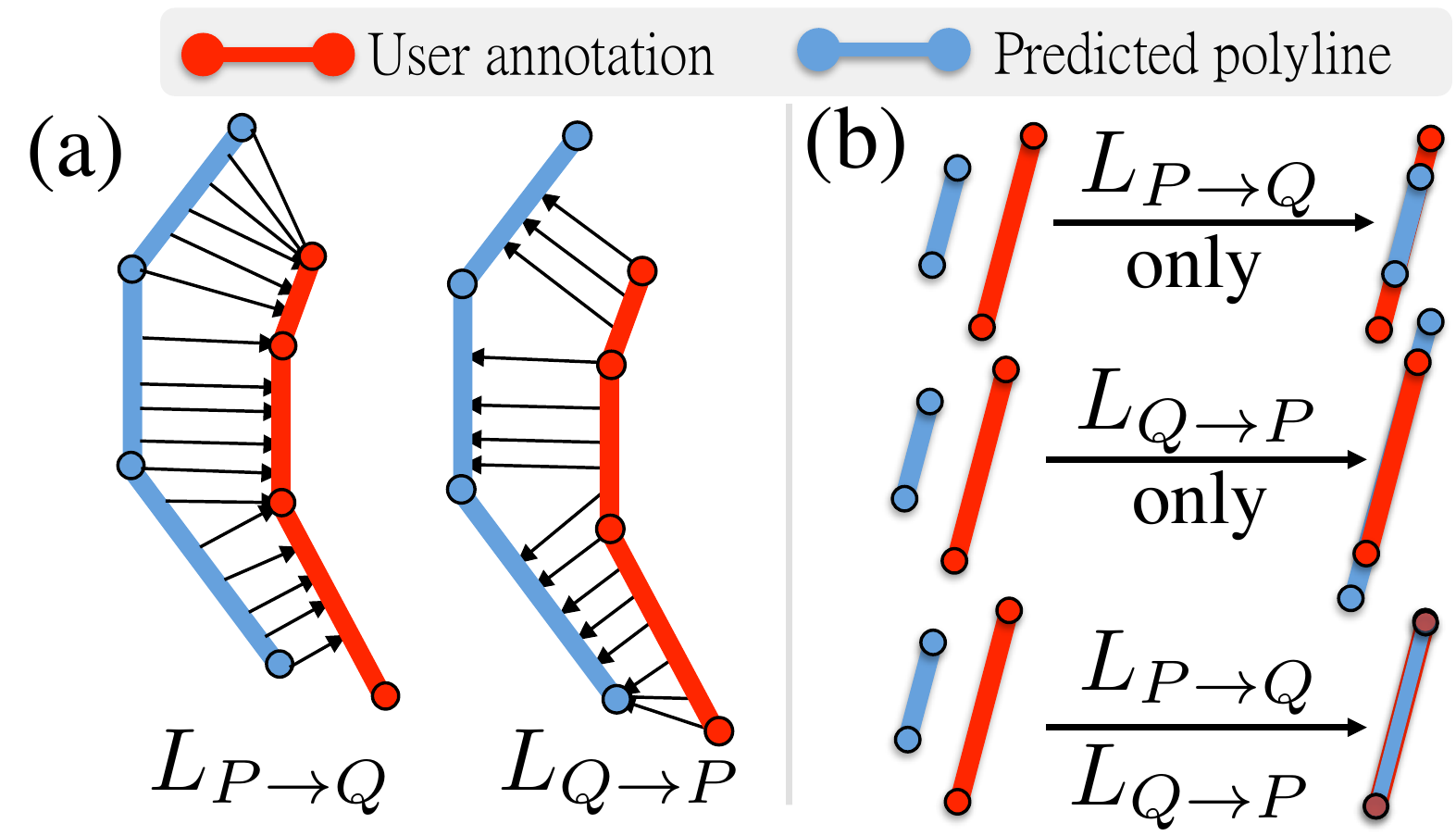}
\end{center}
\caption{(a) Illustration of the two objectives $L_{P\to Q}$ and $L_{Q\to P}$. (b) The effect of the objectives on a toy example. $L_{P\to Q}$ and $L_{Q\to P}$ both have blind spots. By combining both, the model can learn to superimpose perfectly.}
   \label{fig:loss}
   \vspace{-2mm}
\end{figure}

In practice, both loss functions can be obtained by computing the pairwise distances, and then taking a min-pool and finally summing. We illustrate the two terms $L_{P\to Q}$ and $L_{Q\to P}$ in Fig. \ref{fig:loss}(a) and show their effect through a toy example in Fig. \ref{fig:loss}(b).

\begin{figure}[tb]
\begin{center}
   \includegraphics[width=\linewidth]{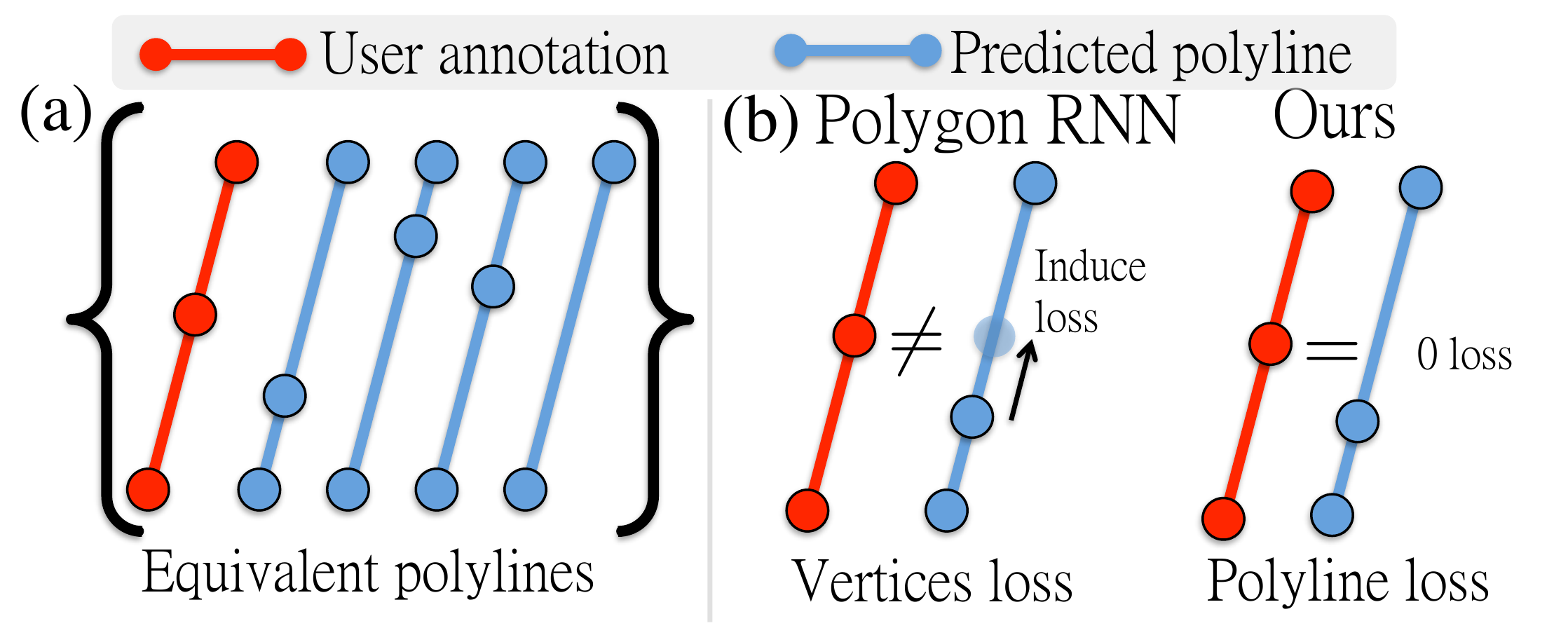}
\end{center}
\caption{(a) A subset of polylines that are equivalent. (b) Comparison to Polygon-RNN \cite{CastrejonCVPR17}. Our loss function encourages the model to learn to draw equivalent polylines, rather than output the exact vertex coordinate.}
   \label{fig:polygon-rnn}
   \vspace{-2mm}
\end{figure}

\vspace{-2mm}
\paragraph{Comparison against Polygon-RNN \cite{CastrejonCVPR17}} 
While our work is inspired by \cite{CastrejonCVPR17}, there exists a critical difference ---
our loss functions are defined w.r.t. the edges rather than the vertices. As shown in Fig. \ref{fig:polygon-rnn}(a), there exist many ways to draw equivalent polylines. It is thus more suitable to consider the distance between polylines than the deviation of the vertices. Fig. \ref{fig:polygon-rnn}(b) shows the caveats of \cite{CastrejonCVPR17}. The prediction can be superimposed perfectly with the ground truth, yet Polygon-RNN still penalizes the model. 
Since polygons are simply special cases of polylines, our polyline loss can be directly plugged into \cite{CastrejonCVPR17}. It can also be applied to other tasks that require learning boundaries, such as boundary detection \cite{yu2017casenet}. We leave this for future study.

\subsection{Attention Loss}

To train the recurrent attention network for polyline counting, we apply a cross entropy loss on the region softmax output $s_t$ and a binary cross entropy loss on the halting probability $h_t$. The ground truth for the regions are the bins  in $x$ where the initial vertex of the lane boundaries falls upon. We present the ground truth bins to the loss function in  order from the left of the image to the right similar to how an annotator picks the initial regions. For the binary cross entropy, the ground truth is equal to one for each lane boundary and  zero when it is time to stop counting.

\subsection{Training Procedure}
We train our model in two stages. At first, the encoder-decoder model with only the polyline-RNN is trained with the ground truth initial regions. We clip the gradients of the conv-lstm to the range of $[-10, 10]$ to remedy the exploding/vanishing gradient problem. For training the conv-lstm, we crop the next region using the predicted previous vertex. The conv-lstm iterates until the next region falls outside the image boundaries or a maximum of image height divided by crop height plus 3. We let the size of the crop to be $60\times 60$ pixels. We train using SGD \cite{bottou2010large} with initial learing rate of 0.001, weight decay 0.0005 and momentum 0.9 for one epoch with a minibatch size of 1. 

Next, we freeze the weights of the encoder and train only the parameters of the recurrent attention module for counting for one epoch. We train the conv-rnn that predicts the number of lane boundaires using the Adam optimizer \cite{adam} with an initial learning rate of 0.0005 and weight decay of 0.0005 with a minibatch size of 20.    
The training criteria were determined based on the results on the validation set. 
The model is trained on one Titan XP GPU for close to 24 hours with the majority of the training time devoted to the Conv-LSTM module.

%!TEX root = ./top.tex

\section{Experimental Evaluation}

\vspace{-2mm}
\paragraph{Dataset:}

We curated a dataset on highways and  mapped a stretch of 90 km and geofenced to rotating consecutive stretches of 10 km for each of training, validation and the test set. Our autonomous vehicle uses a mounted Lidar that captures point clouds at 10 fps.  We sampled uniformly 50,000 frames for the training set and 10,000 frames for each of the validation and the test sets from the corresponding regions. Our data contains both night and day scenes.

For each frame, we project the 3D point cloud and the ground truth lane graph to BEV such that the autonomous vehicle is positioned on the bottom center of the image looking up. We rasterize the lidar point cloud such that each pixel corresponds to 5 cm. We use images of size $960\times 960$ pixels corresponding to 48 m in front and 24 meters on each side of the autonomous vehicle.

\begin{figure}[t]
\[\arraycolsep=1.0pt
\begin{array}{lll}

\includegraphics[width=0.33\linewidth]{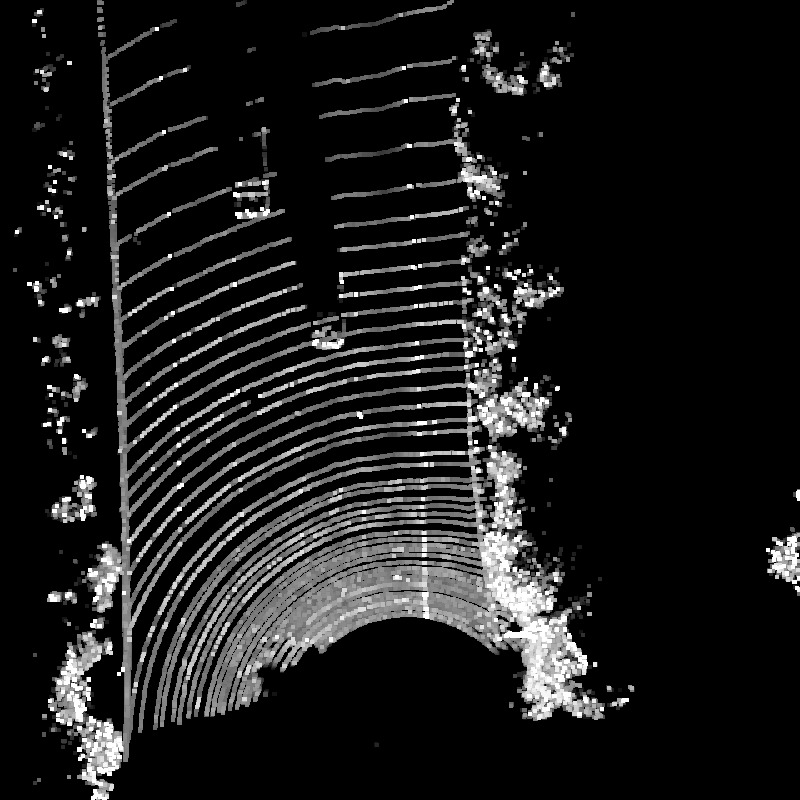}&
\includegraphics[width=0.33\linewidth]{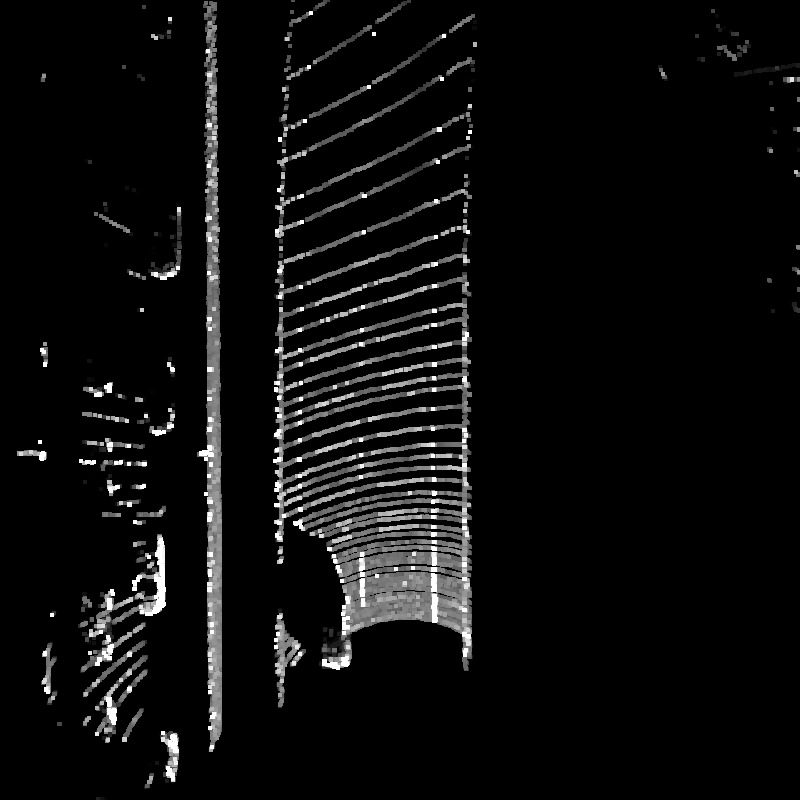}&
\includegraphics[width=0.33\linewidth]{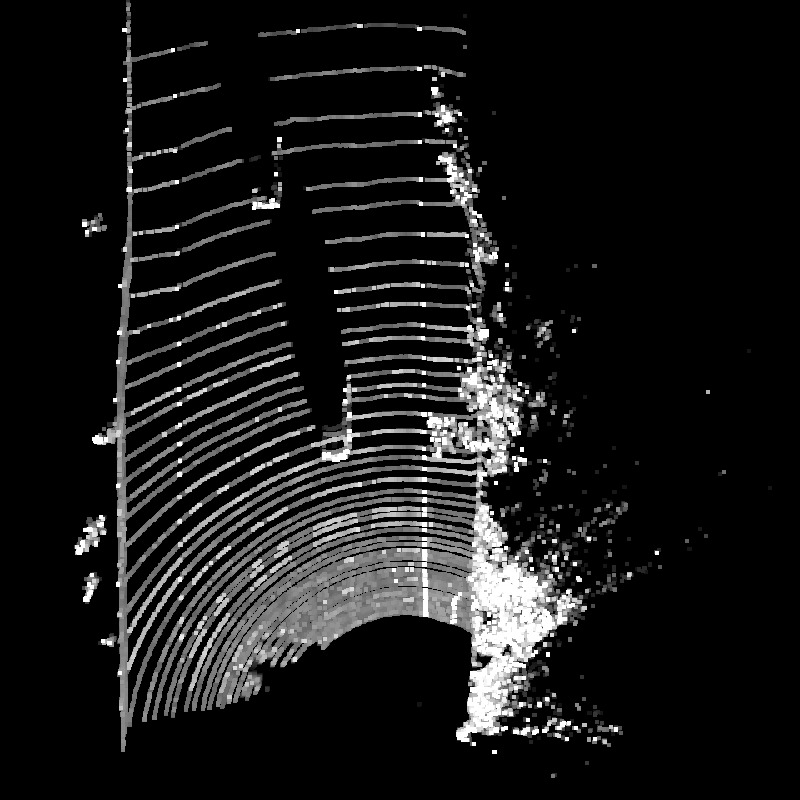}\\

\includegraphics[width=0.33\linewidth]{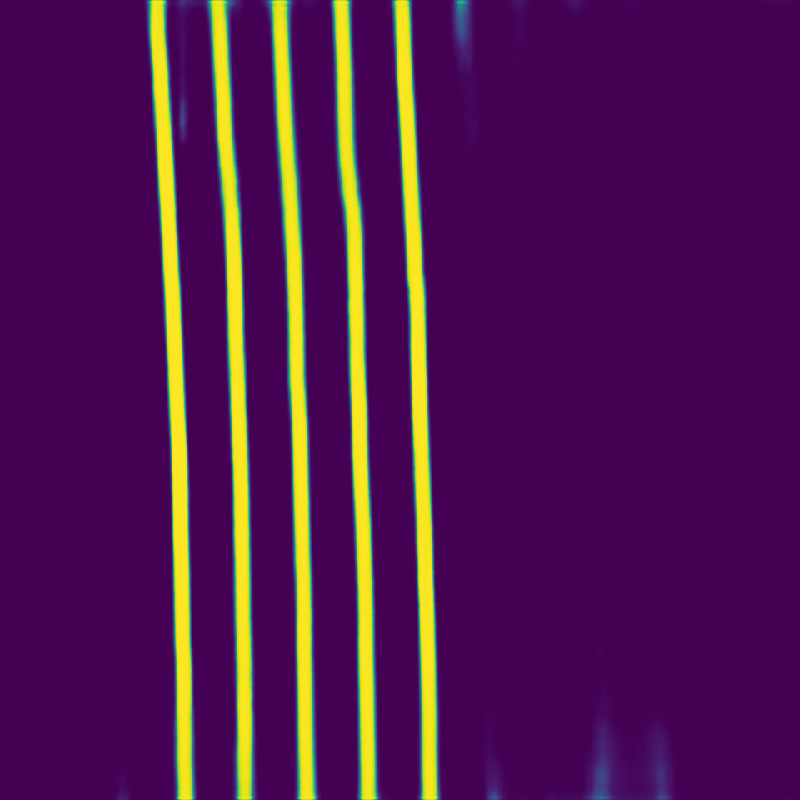}&
\includegraphics[width=0.33\linewidth]{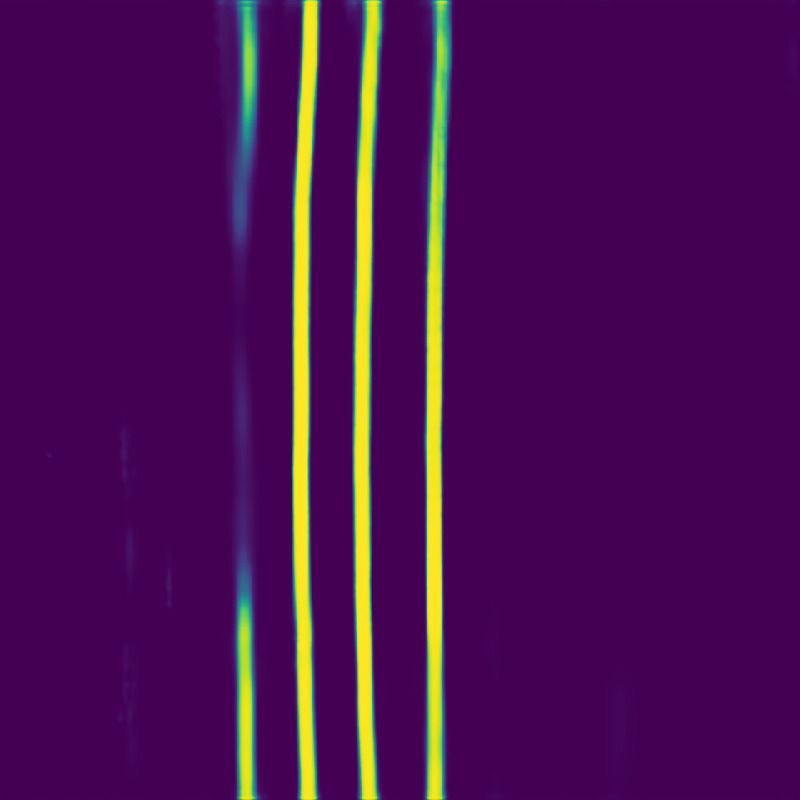}&
\includegraphics[width=0.33\linewidth]{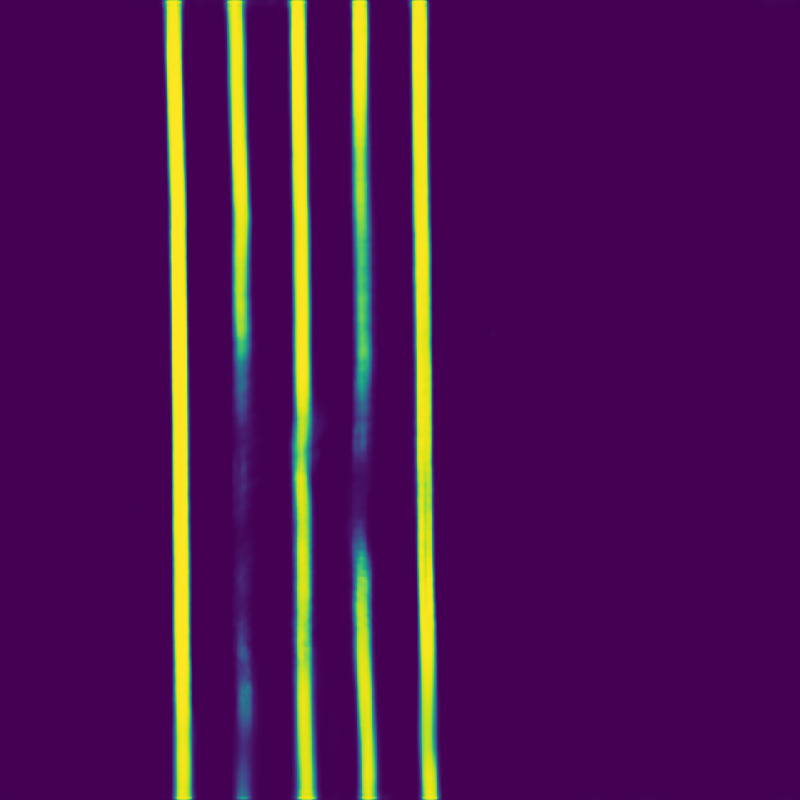}\\

\includegraphics[width=0.33\linewidth]{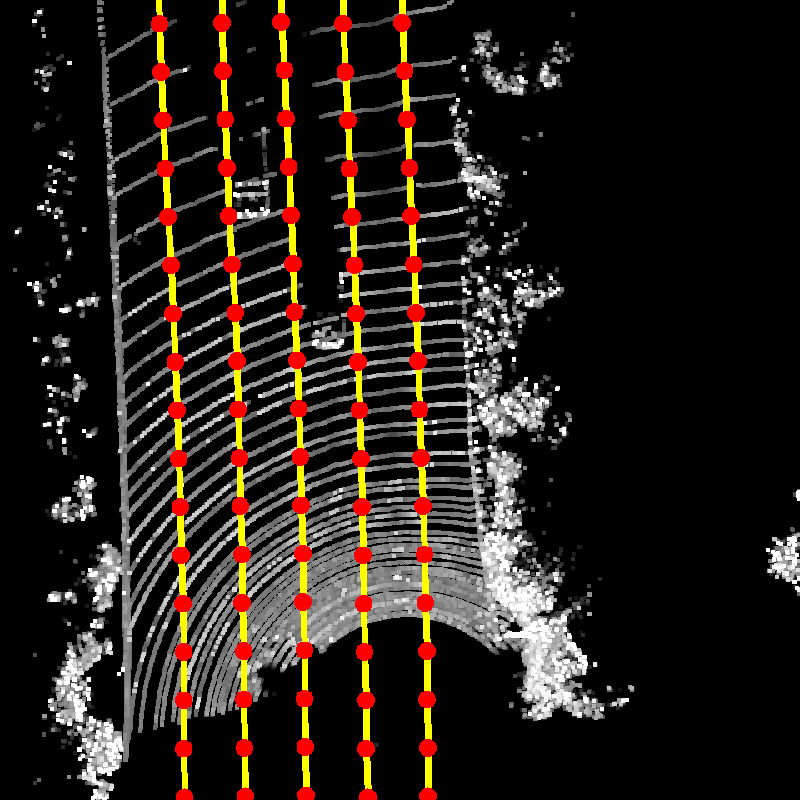}& 
\includegraphics[width=0.33\linewidth]{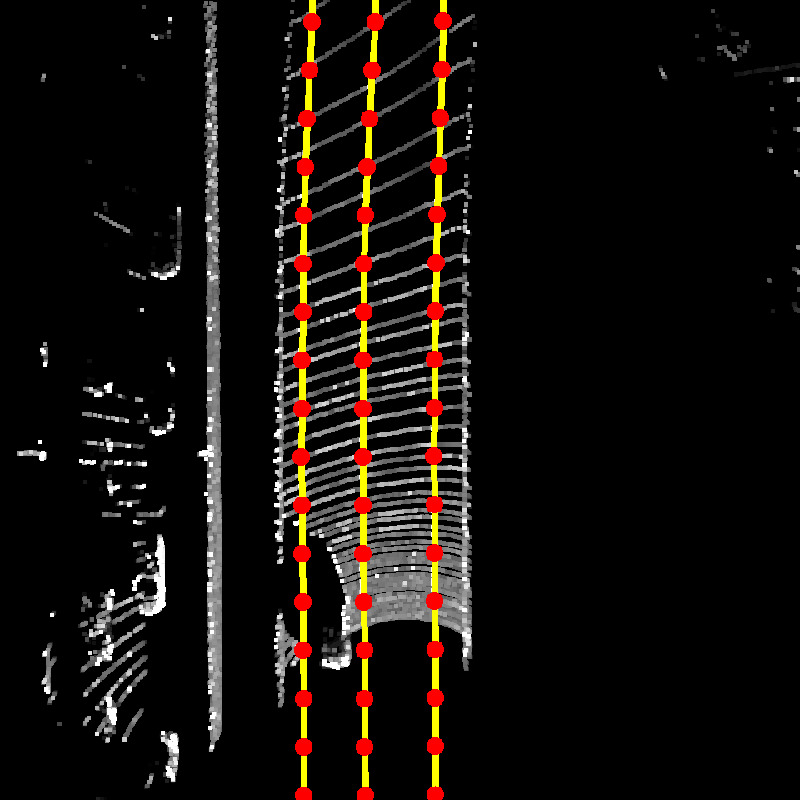}& 
\includegraphics[width=0.33\linewidth]{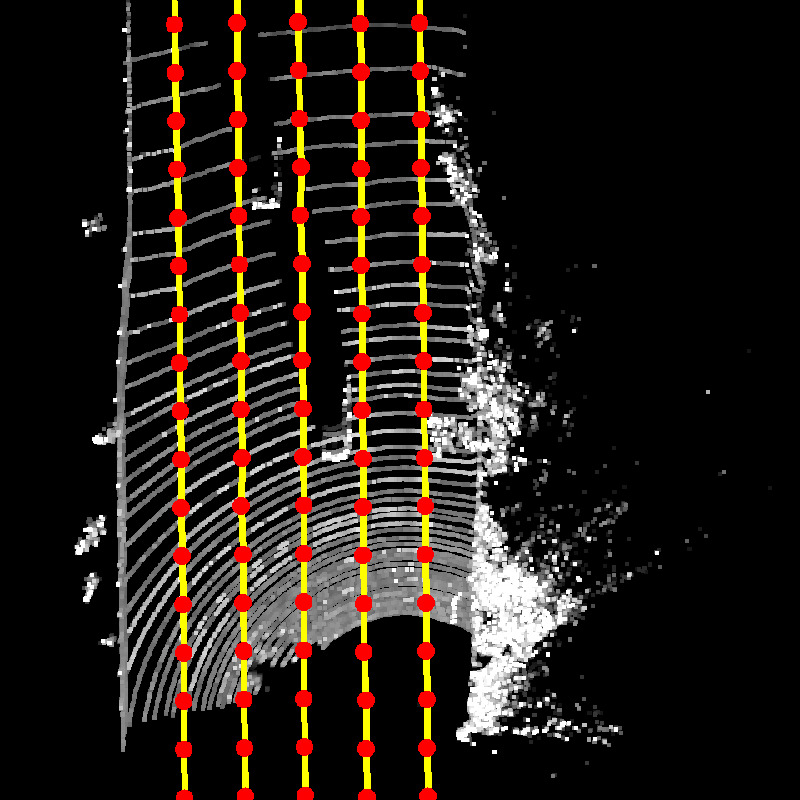}\\

\end{array}
\]
\label{fig:baseline}
\caption{\textbf{Top Row}: Point cloud sweep of the road. \textbf{Middle Row}: The sigmoid output of the baseline corresponding to a 20 pixels wide region around the lane boundaries. \textbf{Bottom Row}: The lane boundary instances outputs of our network.}
\end{figure}

\vspace{-2mm}
\paragraph{Baselines:} Since there are no existing baselines in the literature for this task, we developed a strong baseline to evaluate and motivate our method. In particular, we take the encoder and decoder modules of our architecture, remove the lane counting and Polyline-Rnn modules, and output a sigmoid function corresponding to a dense 20 pixel wide region around each lane boundary. In other words, we aim to detect a dense representation of the lane boundaries. 
We used the exact same architecture of our hierarchical network and trained the network using binary cross-entropy for three epochs. We use the Adam optimizer \cite{adam} with an initial learning rate of 0.001 and weight decay of 0.0001 determined from the validation set. We have visualized some lane detection results in Fig. \ref{fig:baseline}.  

Note that while the output of our hierarchical model is a structured representation where each lane boundary instance is predicted by the network, the baseline only outputs a dense representation of the lane boundaries and further post processing steps are required to obtain each individual instance. Thus, we proceed as follows: First, we threshold the sigmoid output of the baseline for different values of 0.3, 0.5, 0.7 and 0.9 to remove spurious detections. Each threshold is considered as a baseline and we refer to them as CE at 0.3 to Ce at 0. Next, separately for each baseline, we skeletonize the result and finally obtain each individual lane boundary instance using connected components.

\begin{figure}[t]
\begin{center}
   \includegraphics[width=\linewidth]{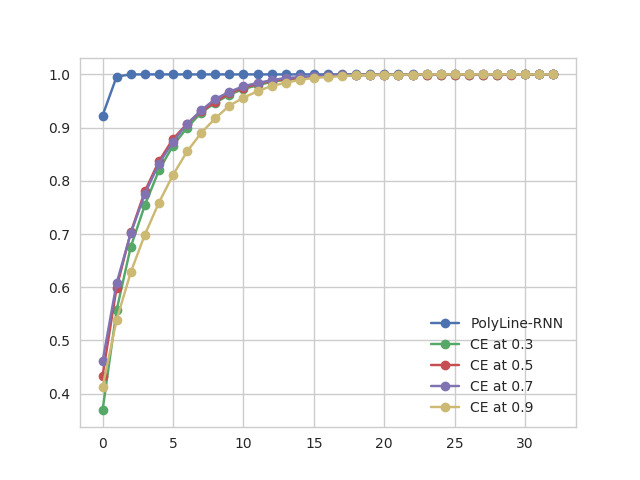}
\end{center}
\caption{The cumulative distribution of the absolute value of the difference between the ground truth number of lanes and the predicted number of lanes by our network as well as the baselines.}
\label{fig:topology}
\vspace{-4mm}
\end{figure}

\vspace{-2mm}
\paragraph{Topology:} Our first metric focuses on evaluating whether the correct number of lanes is estimated. In Fig. \ref{fig:topology}, we demonstrate the cumulative distribution of the absolute value of the difference between the ground truth number of lanes and the predicted number of lanes by our network as well as the baselines. Our model estimates the correct number of lanes 92\% of the time while deviating by one or less lane boundaries almost 99\% of the time. We note that our method outperforms the strongest baseline (in terms of topology) that retrieves the correct number of lanes only 46\% of the time while being one away or less 61\% of the time. We highlight that our model is specifically designed to output a structured representation of the lane boundaries by learning to count and draw polylines. On the other hand, the ad-hoc post processing steps applied to the baseline introduce holes in places where the lane detections do not fire and as such topology deteriorates. Hence the reason some of the baselines predict more than 10 lane boundaries. Moreover, our structured representation enables the annotator to easily correct a mistake by either specifying the initial region of a lane boundary (if it's missed) or by deleting it altogether. This benefit is not extendible to the baselines. 
We will present an experiment later on that corroborates this aspect of our model.

\begin{figure*}[!t]
\label{fig:good}
\[\arraycolsep=1.0pt
\begin{array}{llllll}
\includegraphics[width=0.16\linewidth]{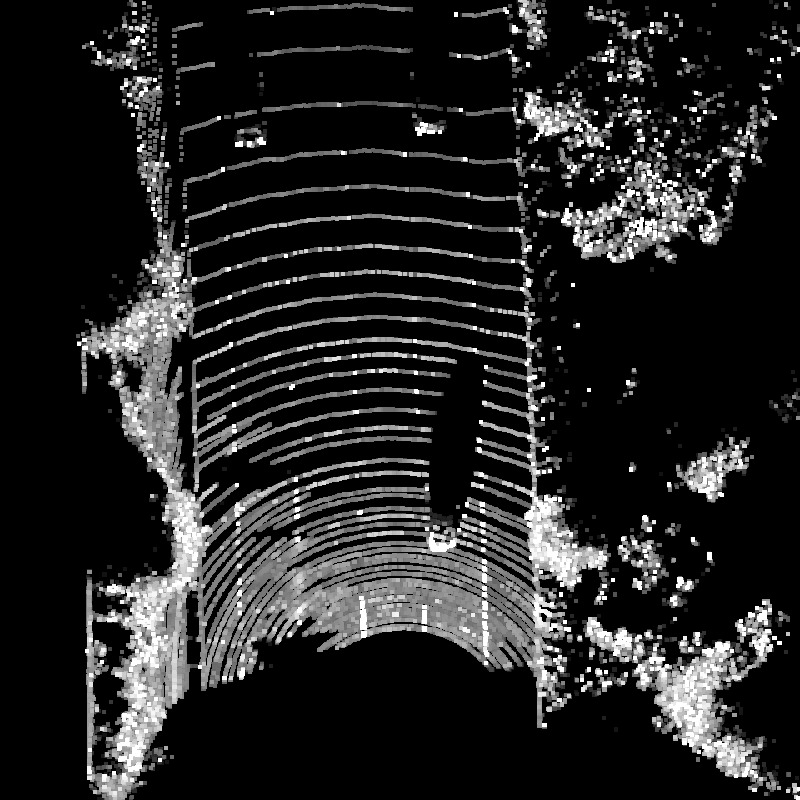}& 
\includegraphics[width=0.16\linewidth]{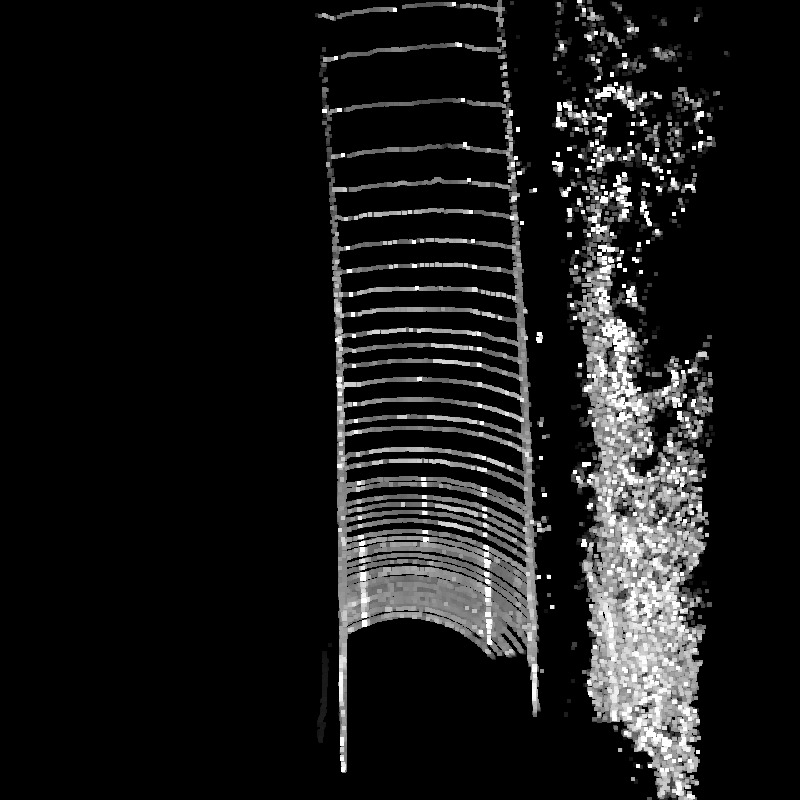}  & 
\includegraphics[width=0.16\linewidth]{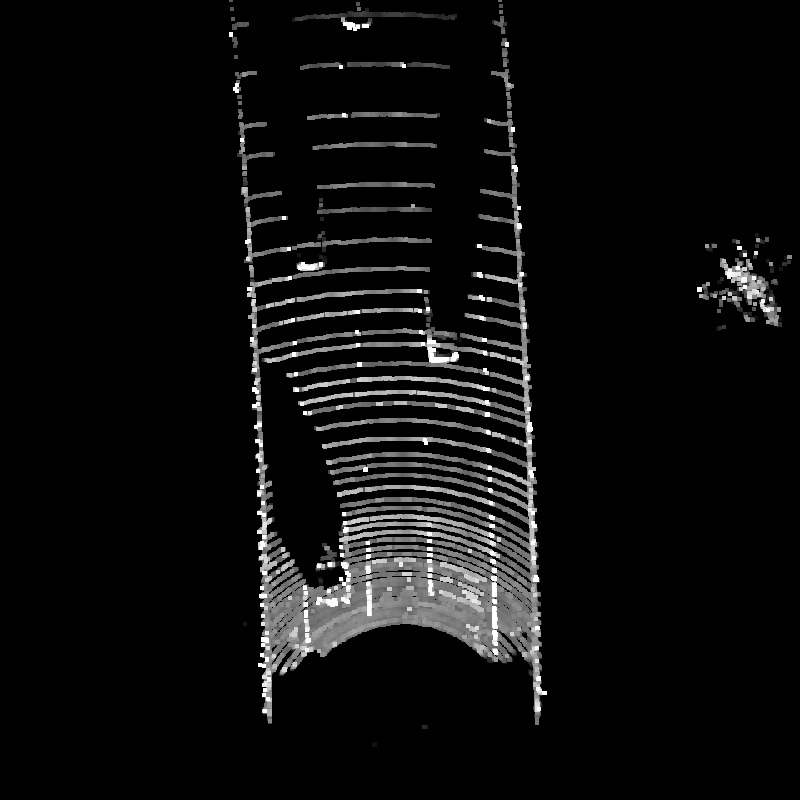}  & 
\includegraphics[width=0.16\linewidth]{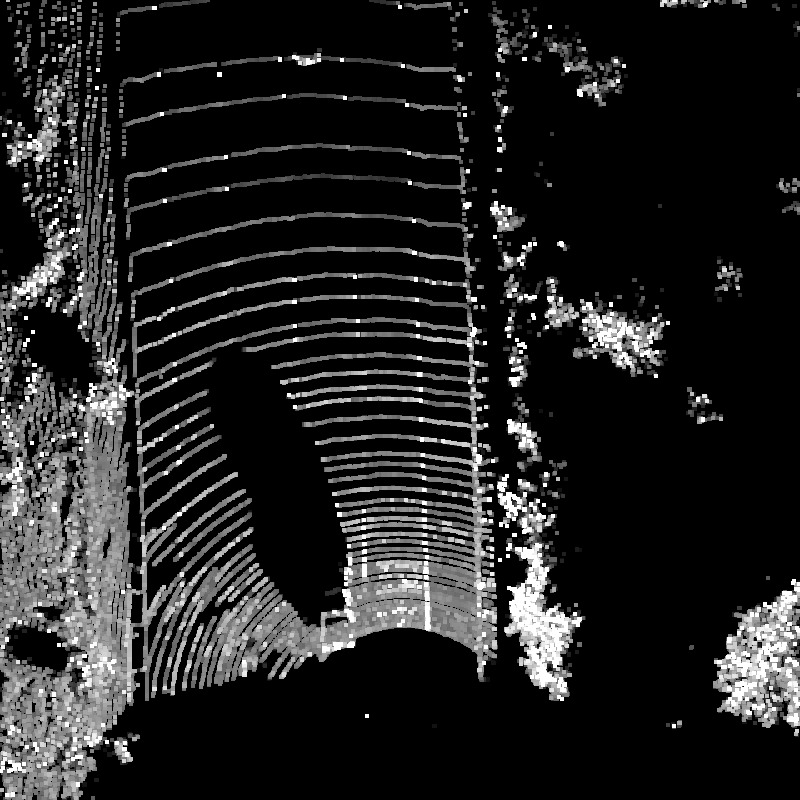}  & 
\includegraphics[width=0.16\linewidth]{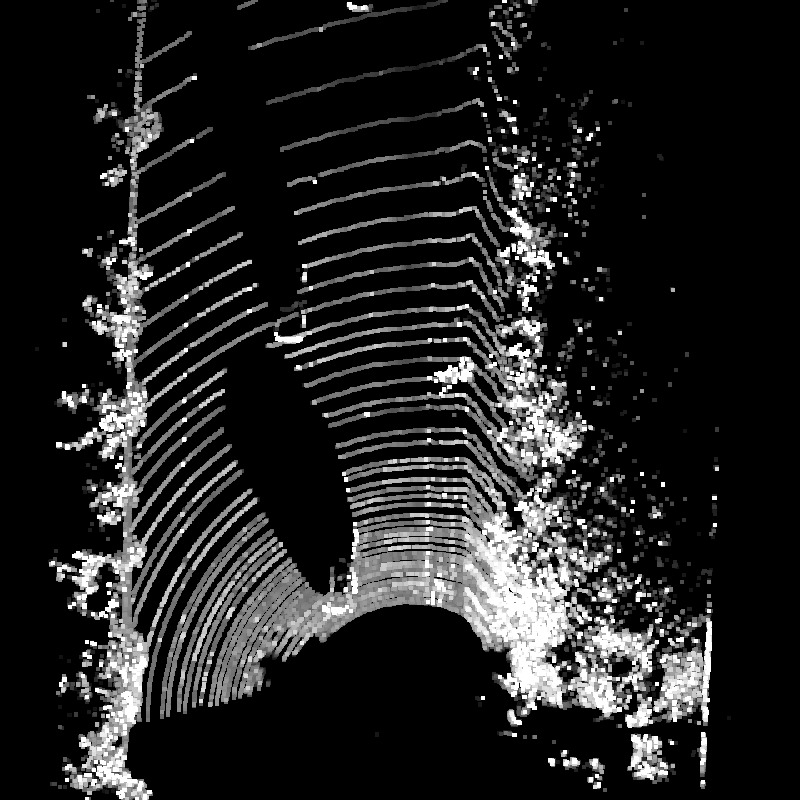}& 
\includegraphics[width=0.16\linewidth]{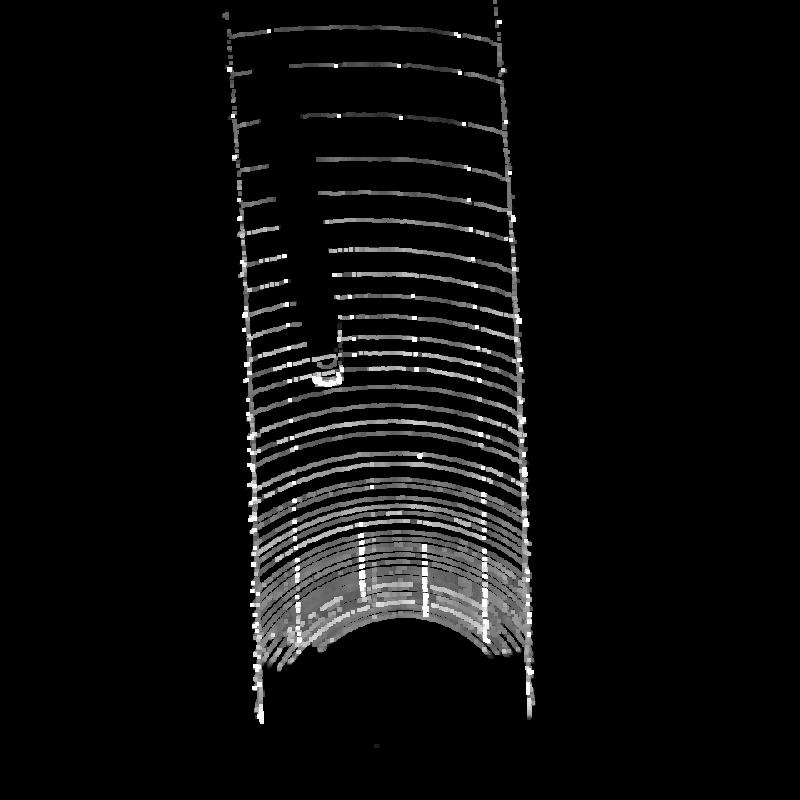}\\

\includegraphics[width=0.16\linewidth]{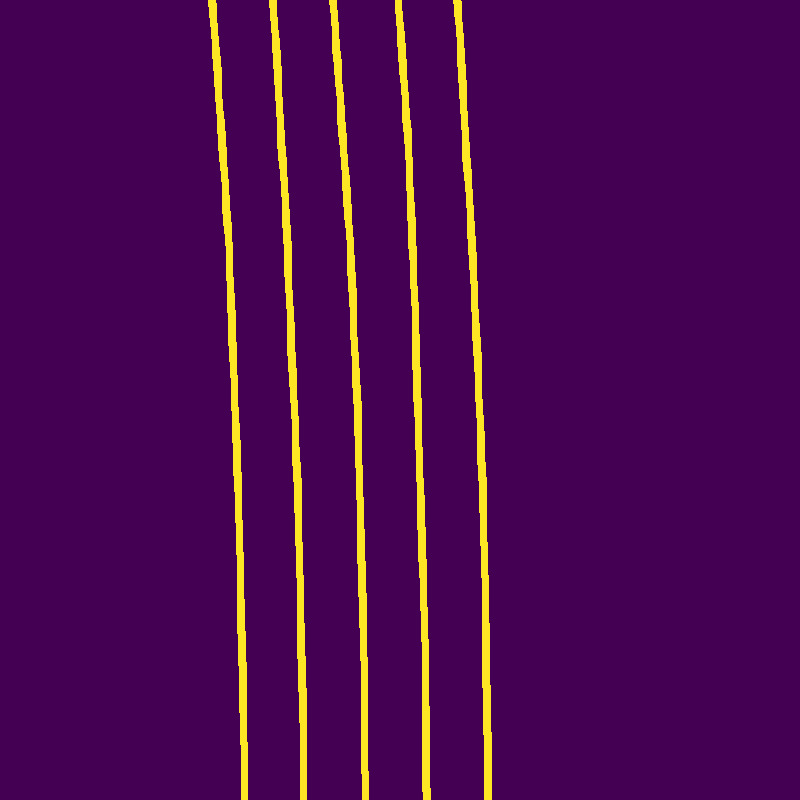}&
\includegraphics[width=0.16\linewidth]{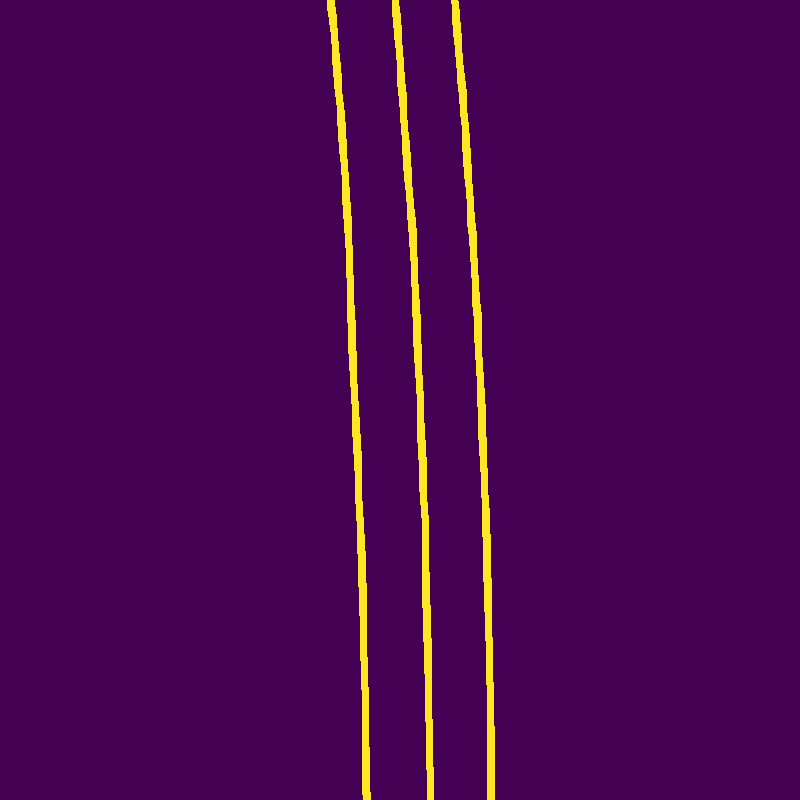}  & 
\includegraphics[width=0.16\linewidth]{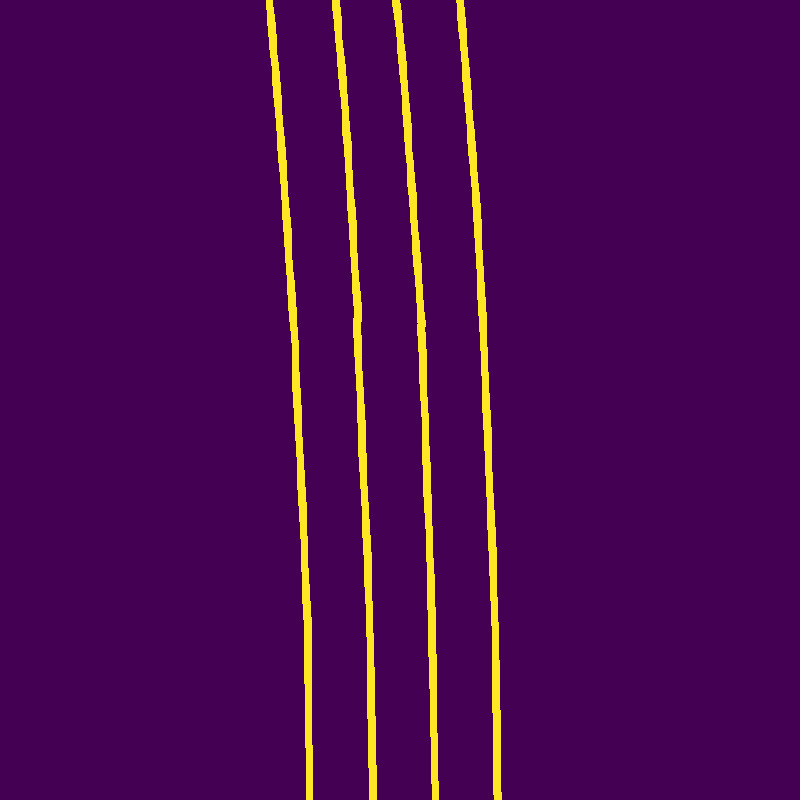}  & 
\includegraphics[width=0.16\linewidth]{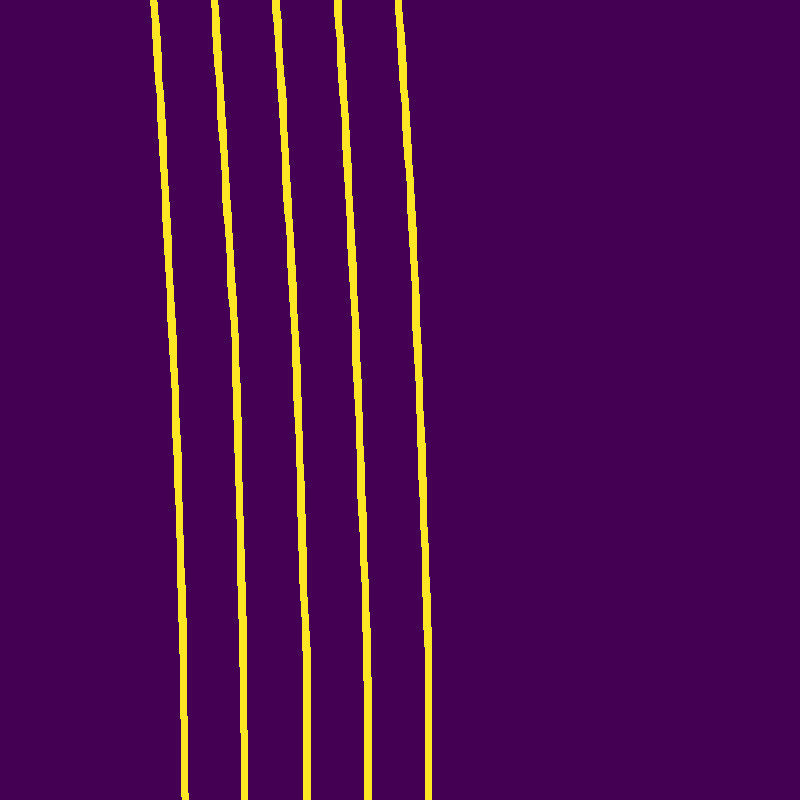}  & 
\includegraphics[width=0.16\linewidth]{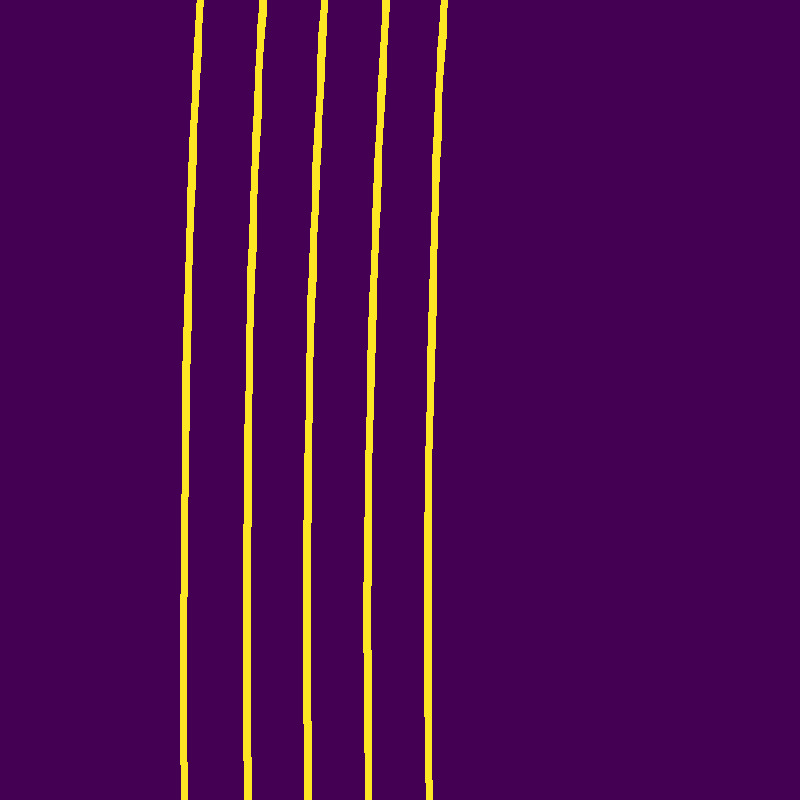}& 
\includegraphics[width=0.16\linewidth]{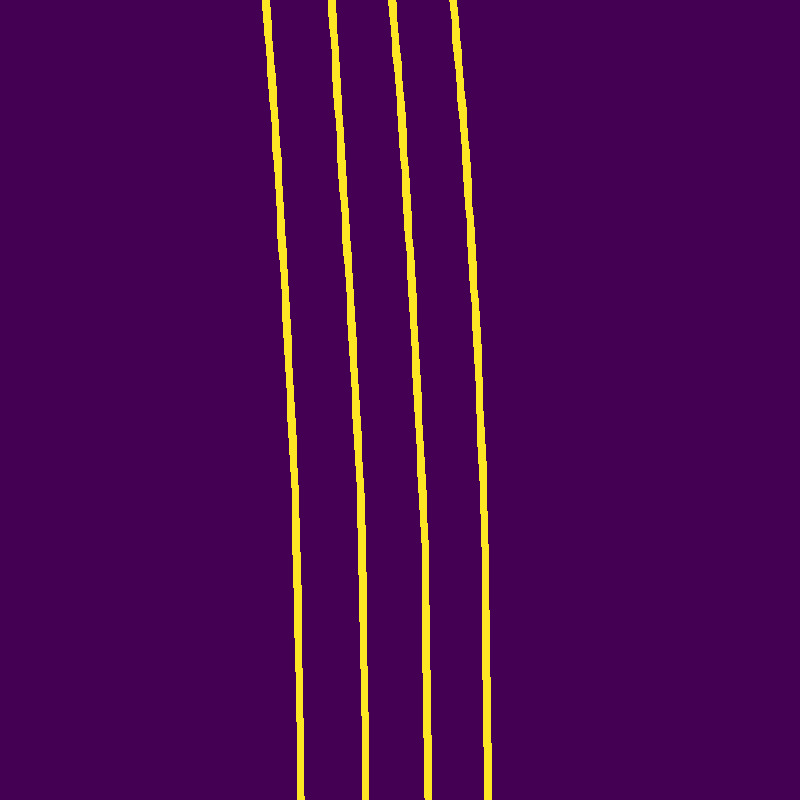}\\

\includegraphics[width=0.16\linewidth]{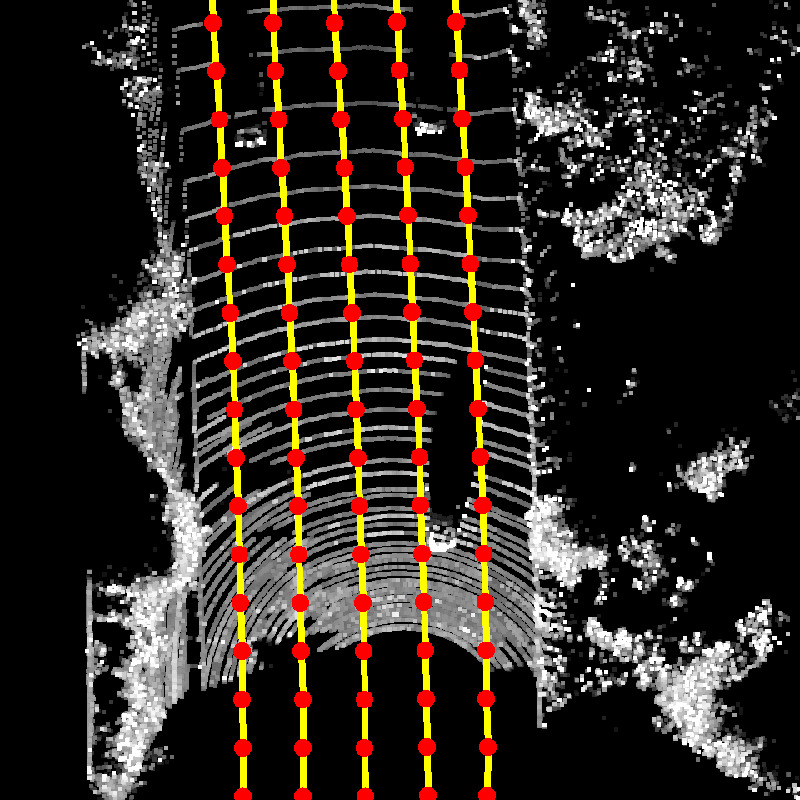}& 
\includegraphics[width=0.16\linewidth]{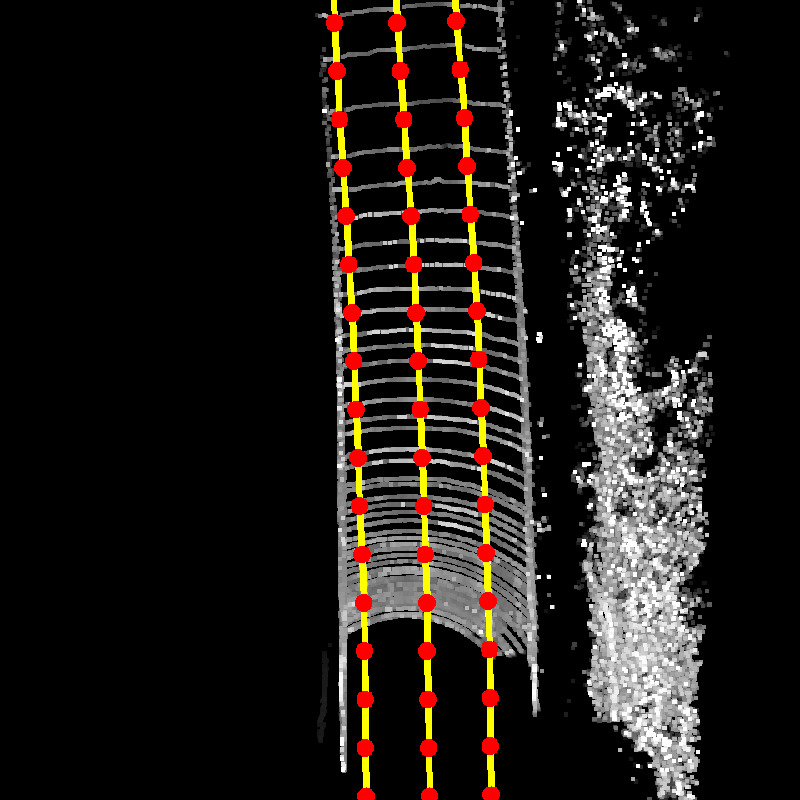}  & 
\includegraphics[width=0.16\linewidth]{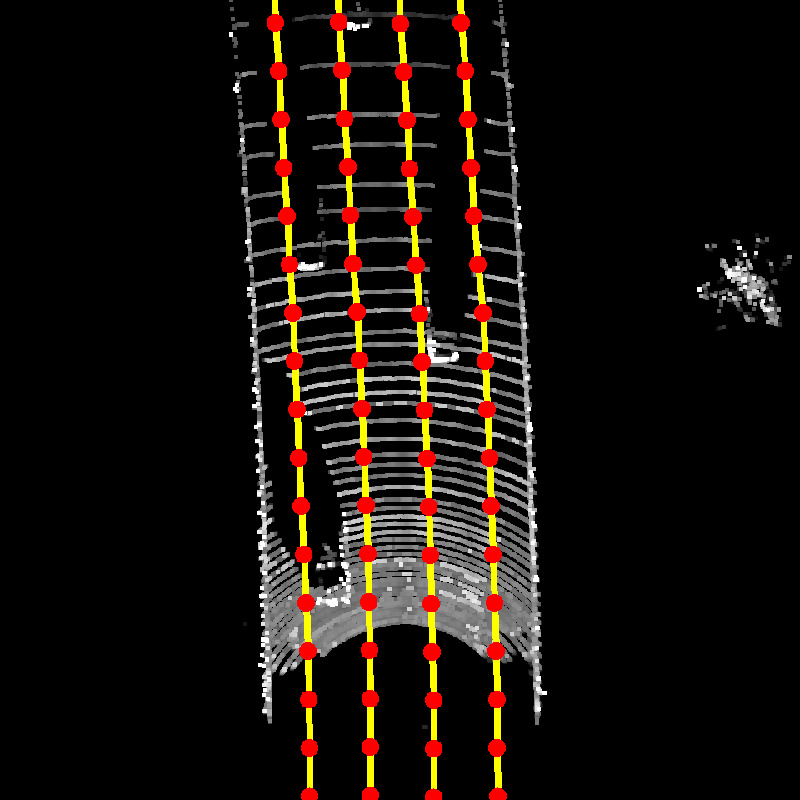}  & 
\includegraphics[width=0.16\linewidth]{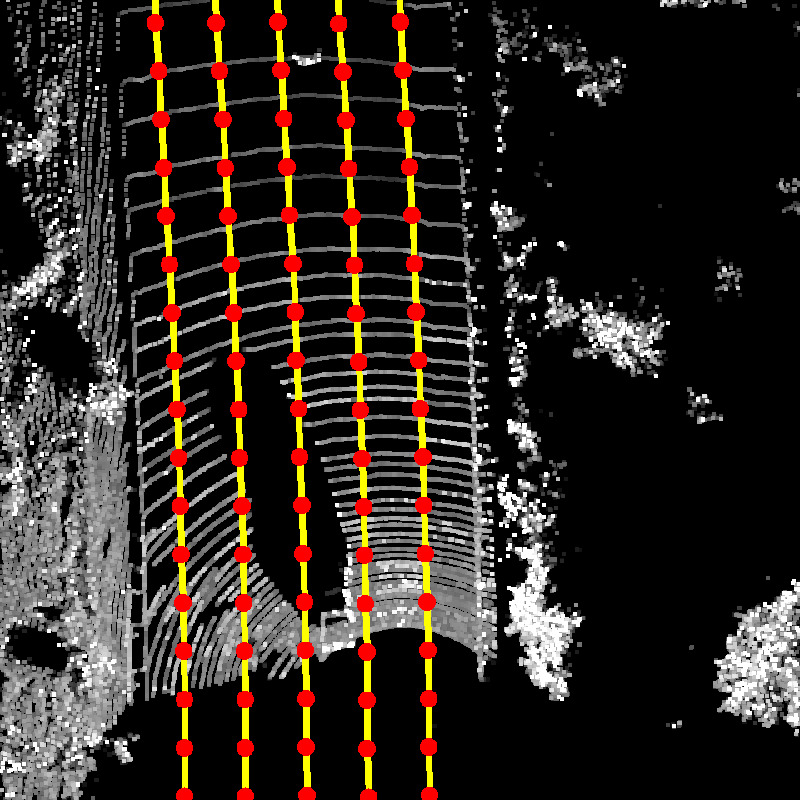}  & 
\includegraphics[width=0.16\linewidth]{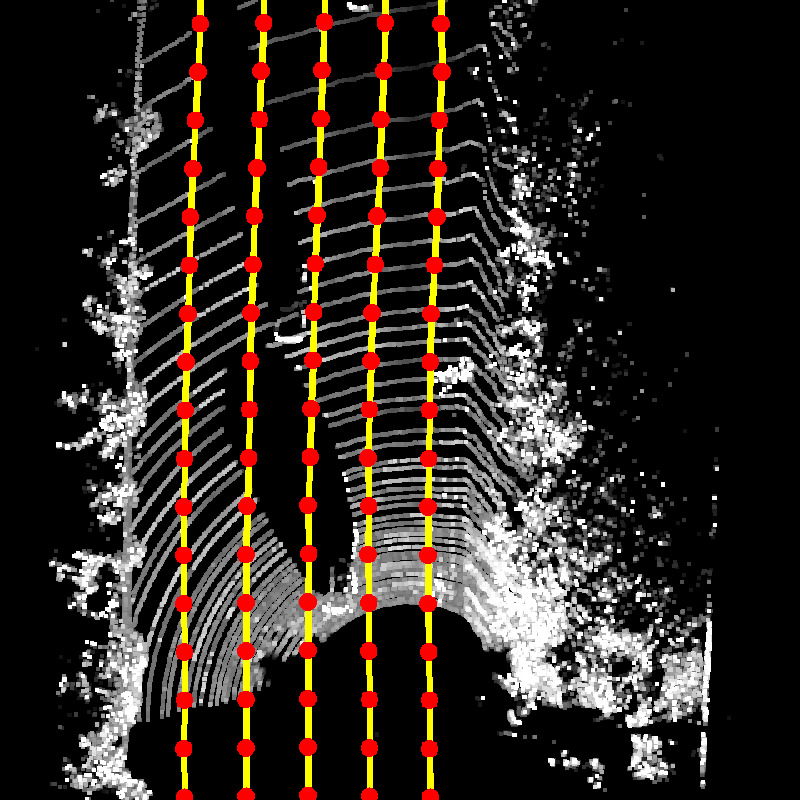}& 
\includegraphics[width=0.16\linewidth]{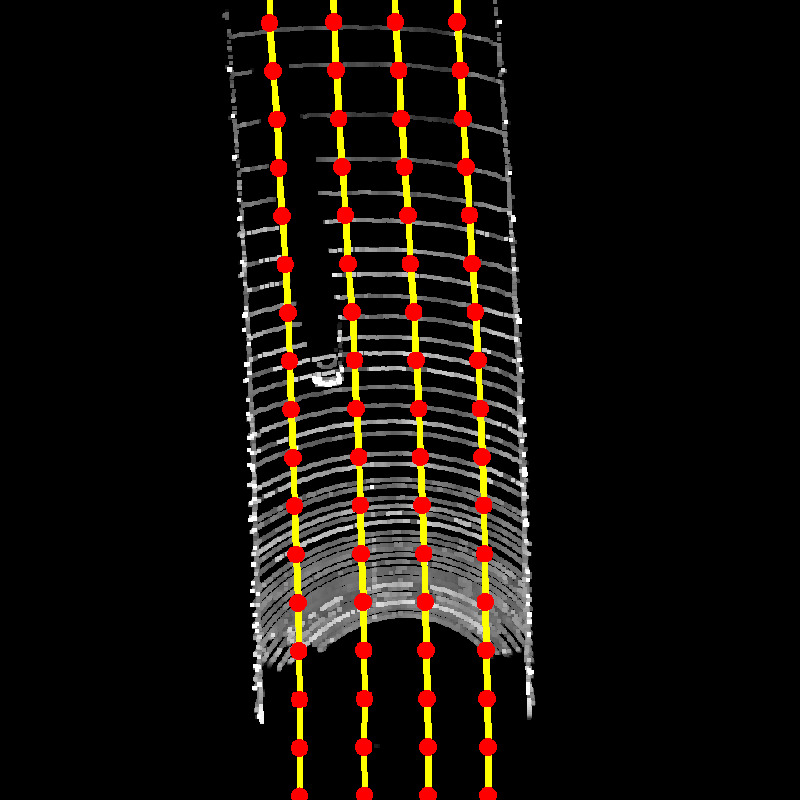} \\

\includegraphics[width=0.16\linewidth, trim={0.8cm 0 0 0}]{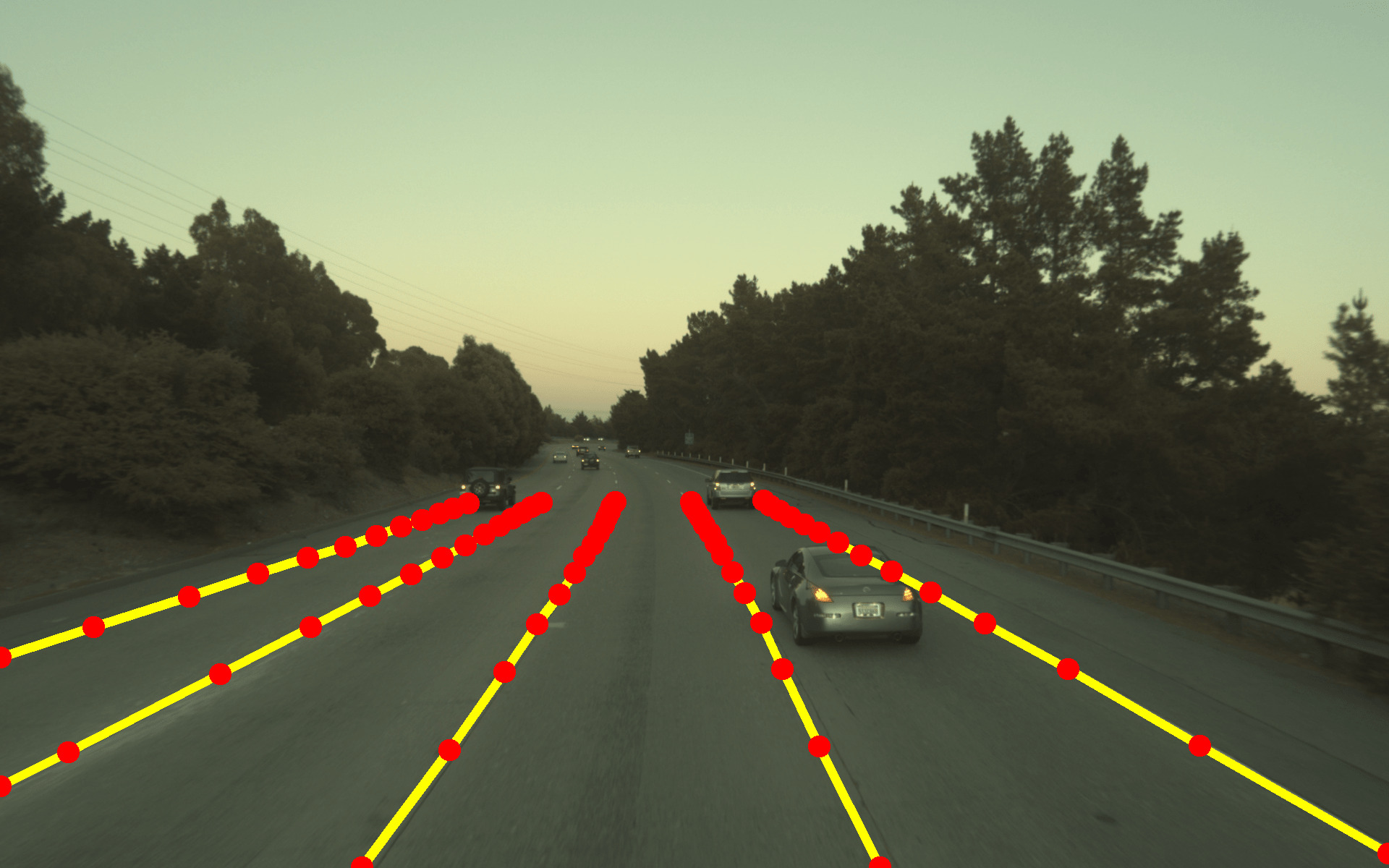}& 
\includegraphics[width=0.16\linewidth, trim={0.8cm 0 0 0}]{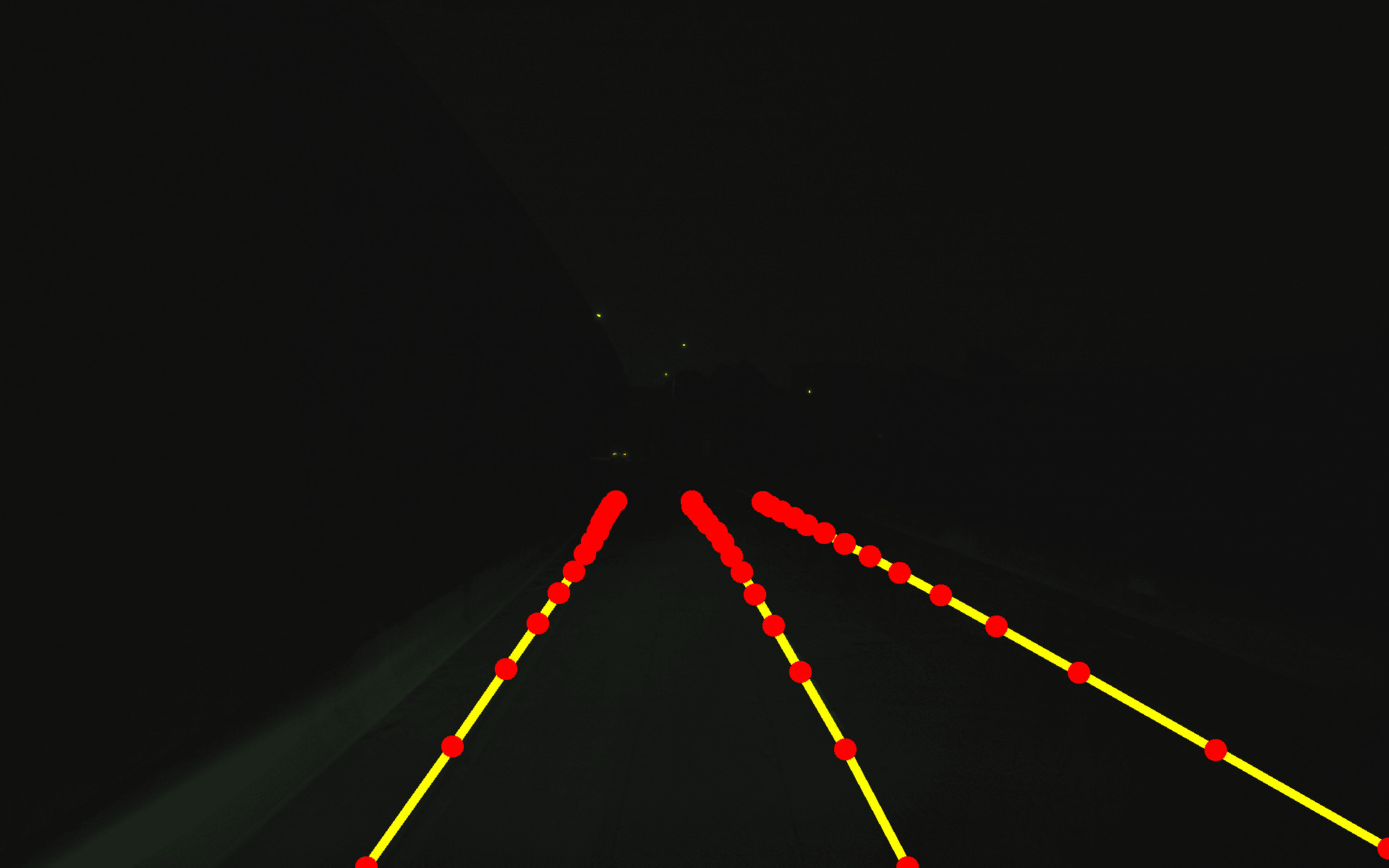}  & 
\includegraphics[width=0.16\linewidth, trim={0.8cm 0 0 0}]{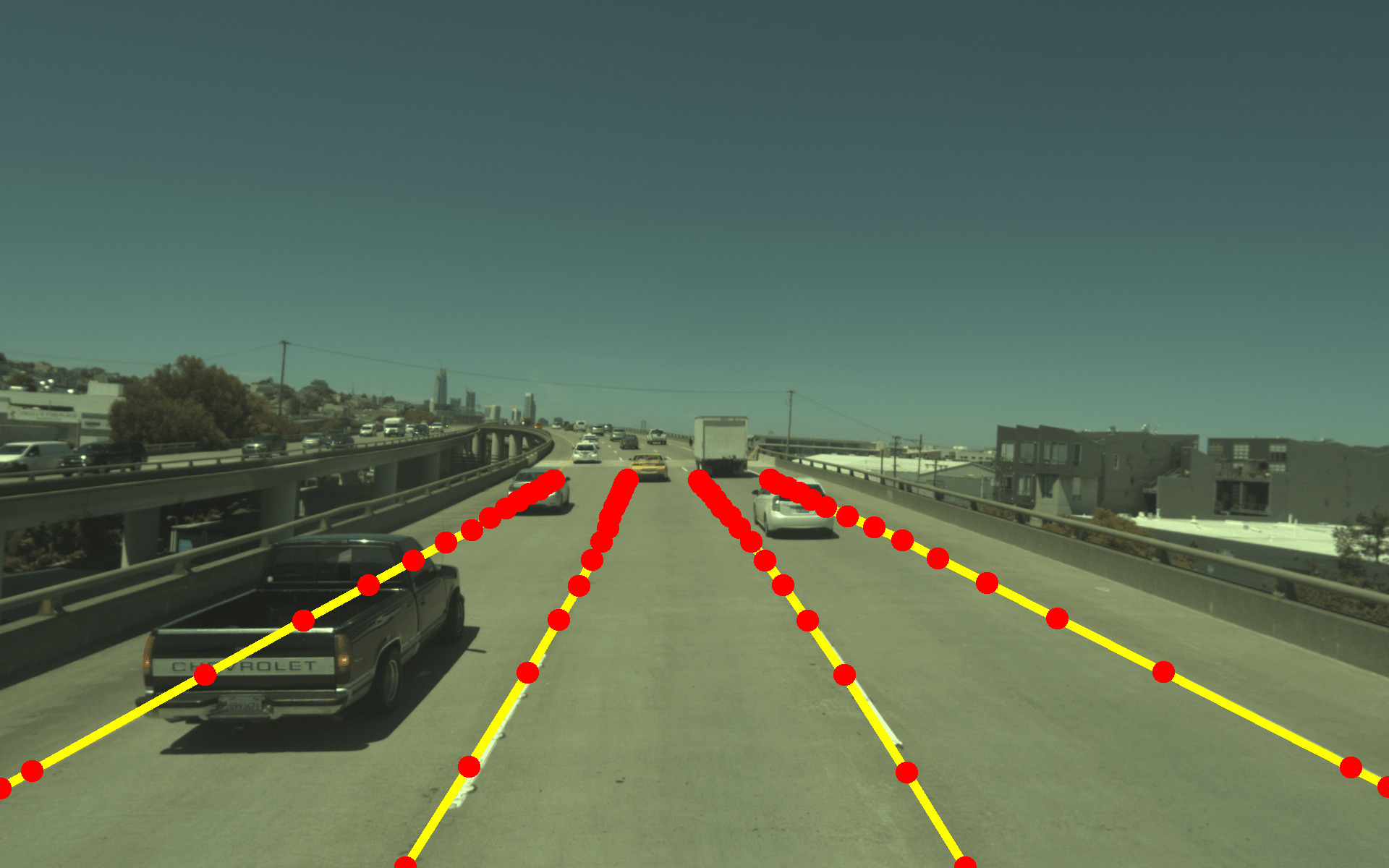}  & 
\includegraphics[width=0.16\linewidth, trim={0.8cm 0 0 0}]{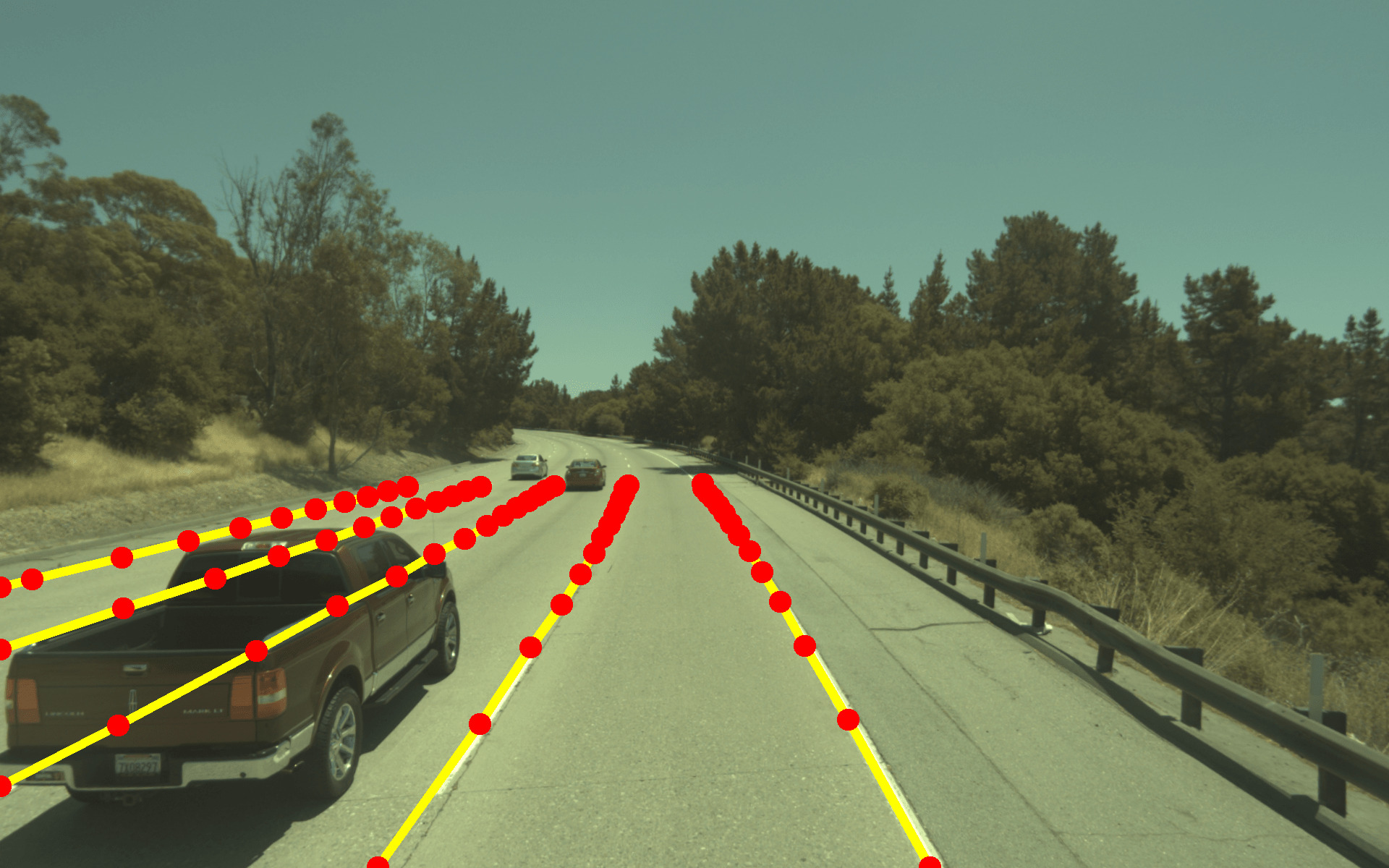}  & 
\includegraphics[width=0.16\linewidth, trim={0.8cm 0 0 0}]{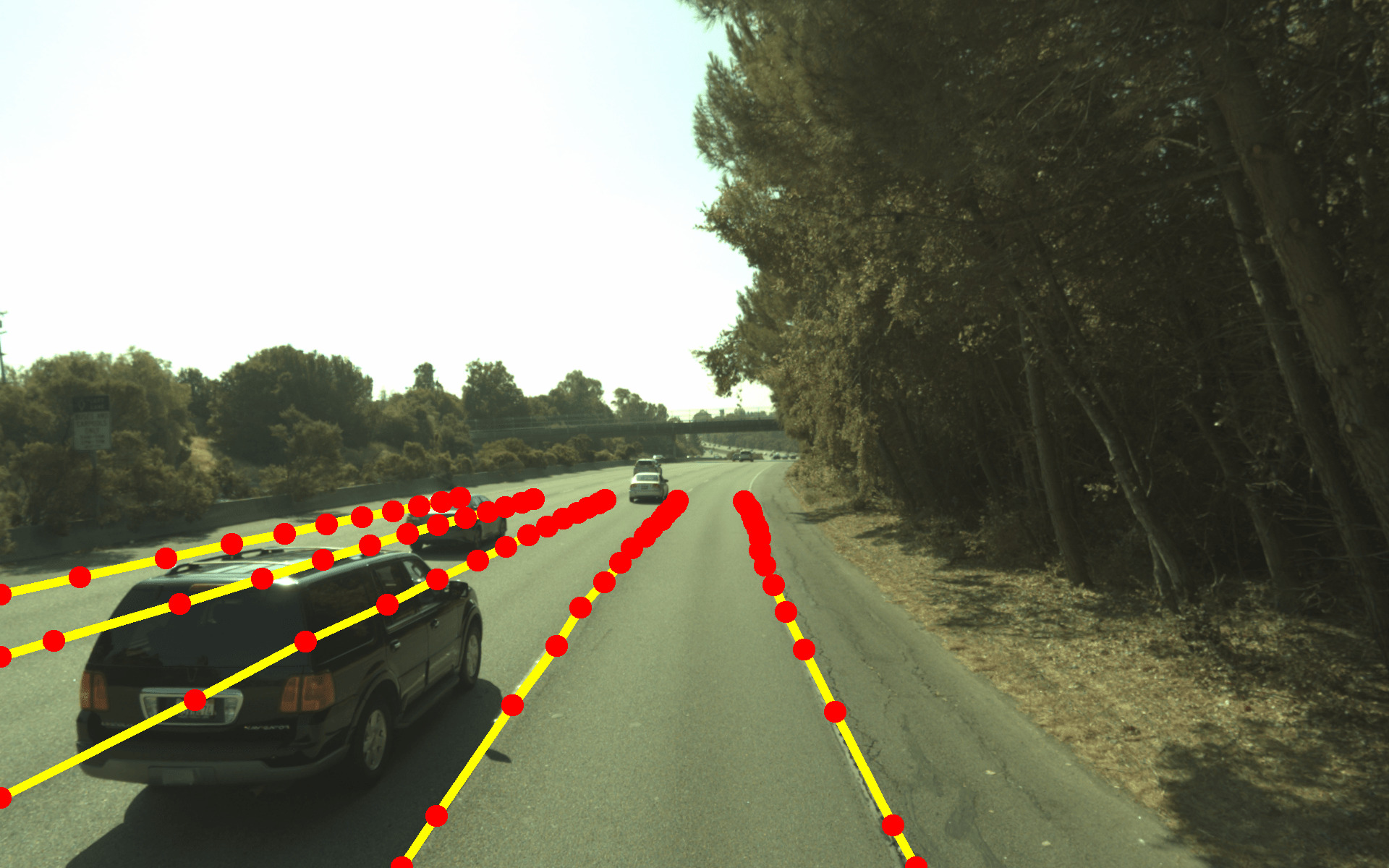}& 
\includegraphics[width=0.16\linewidth, trim={0.8cm 0 0 0}]{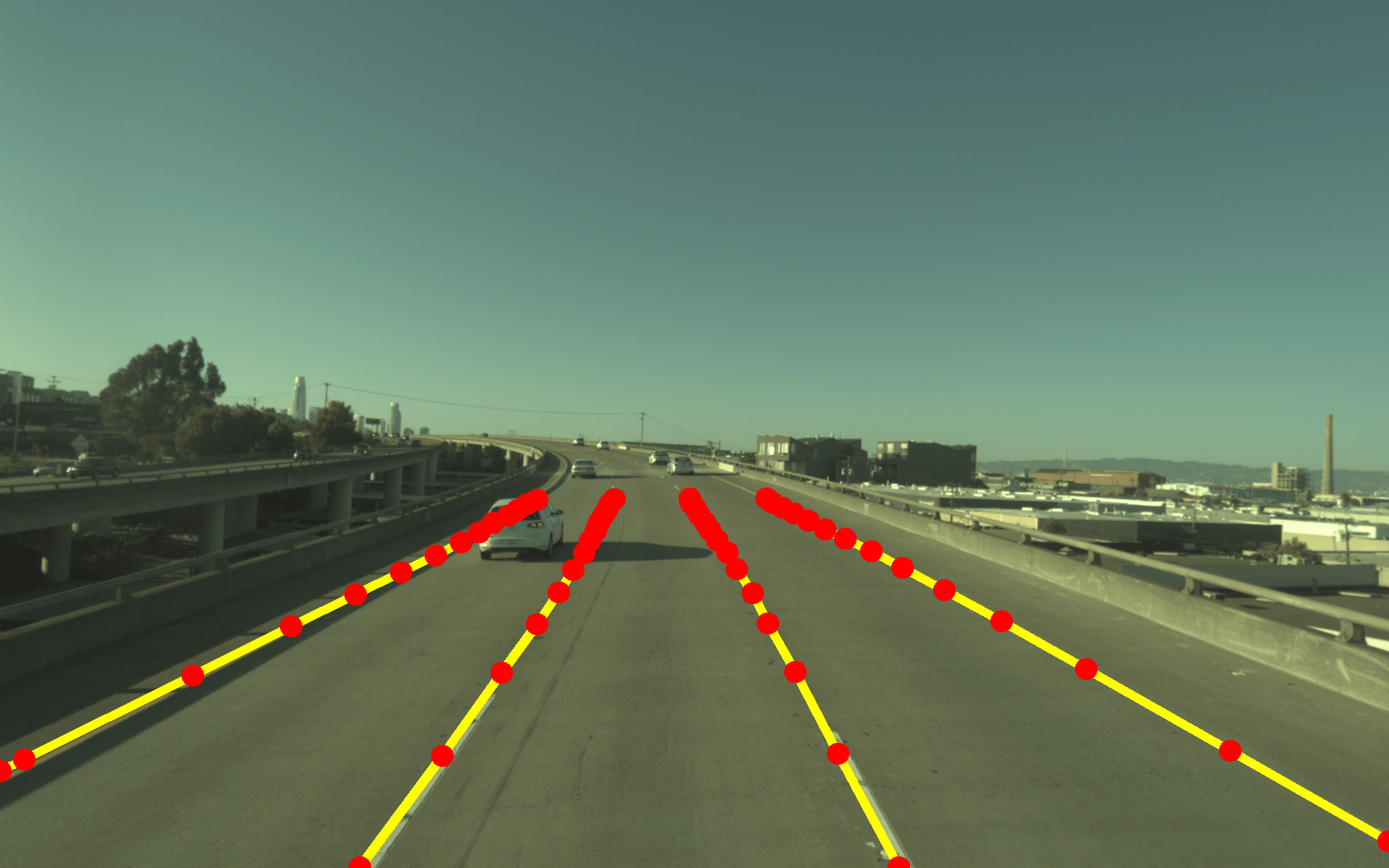}

\end{array}
\]
\caption{Qualitative Examples. \textbf{Top Row}: Point cloud sweep of the road. \textbf{Second Row}: The ground truth lane graph. \textbf{Third Row}: The lane boundary instances outputs of our network. \textbf{Bottom Row}: The predicted lane boundaries projected onto the image.}
\end{figure*}

\vspace{-2mm}
\paragraph{Precission/Recall:}
We use the precision and recall metric of \cite{TCity2017}. We define precision as the number of predicted polyline points within a threshold distance of a ground truth lane boundary divided by the total number of points on the predicted polylines. Recall is defined in a symmetric fashion as with respect to a thresholded distance of the predicted polylines. We set the thresholds at the intervals of 5, 10, 15, and 20 cm corresponding to a maximum deviation of 4 pixels from a polyline. Note that bigger thresholds are not interesting as the autonomous vehicle should be able to localize itself with high accuracy. 

From  Table. \ref{table:prec_recall} one can note that the best performing baseline in terms of precision, i.e. CE at 0.9, performs the worst in comparison to CE at 0.5 that achieves the best recall. This alludes to the fact that when only a small number of points are retained for a higher threshold, precision goes up while the opposite holds for recall. 

Our model beats the baselines in both precision and recall. Although the values are within 1-2 percentage points of each other. We remind the reader that our method beats the best performing baseline with a high margin when it comes to retrieving the correct topology.

\begin{table*}[]
\label{table:prec_recall}
\centering

\begin{tabular}{c|c|l|c|c||c|l|l|l|}
\cline{2-9} & \multicolumn{4}{c||}{Precision at (cm)} & \multicolumn{4}{c|}{Recall at (cm)} \\ \cline{2-9} 
 & 5   & 10   & 15& 20   & 5  & 10   & 15   & 20   \\ \hline

\multicolumn{1}{|c|}{CE at 0.3} &  0.200  & 0.550   & 0.773    & 0.879  & 0.203   & 0.560  & 0.788 & 0.896 \\ \hline

\multicolumn{1}{|c|}{CE at 0.5} &  0.211   & 0.575   & 0.796  & 0.896   & 0.209     & 0.574  & 0.799 & 0.894 \\ \hline

\multicolumn{1}{|c|}{CE at 0.7} &   0.212 & 0.577   & 0.801   & 0.903    & 0.207   & 0.566  & 0.787 & 0.887 \\ \hline

\multicolumn{1}{|c|}{CE at 0.9}  &  0.212  & 0.580   & 0.810   & 0.917    & 0.198    & 0.546  & 0.762 & 0.861 \\ \hline

\multicolumn{1}{|c|}{Ours}  & \textbf{0.226}  & \textbf{0.609}  &  \textbf{0.827} & \textbf{0.92 } & \textbf{ 0.223} & \textbf{0.6}   & \textbf{0.816}    & \textbf{0.908} \\ \hline
\end{tabular}

\caption{Comparison of our proposed model vs. the cross entropy baseline in terms of precision and recall for distances of 5 to 20 cm from the lane boundaries.}
\vspace{-4mm}

\end{table*}

\vspace{-2mm}
\paragraph{Annotator In the Loop:}

The output of our model is a structured representation of the lane boundary instances and as such is easily amenable to bring an annotator in the loop. To demonstrate this, we perform an experiment on the set of 687 examples where we predict the wrong topology. In particular, the annotator either clicks on the starting region of a lane boundary or removes one by just a click when it is either missed or hallucinated respectively. In Table. \ref{table:annot}, we observe that among these failure cases, precision suffers after correction by a maximum of 2 \% for different distances to the lane boundary while recall increases by at least 1\% and maximum 10\% at a 20 cm distance to the lane boundary. This is expected since there is usually low evidence for a lane boundary in failure cases and adding just the starting region would improve only the recall but have an adverse effect on the overall precision. Importantly, the annotator takes on average 1.07 clicks to fix these issues. 

We highlight that the annotator only needs to specify a coarse starting region of the lane boundary, e.g. Fig. \ref{fig:grid2}, rather than an exact initial vertex. This facilitates the task; One can see from the Lidar images of Fig \ref{fig:good} that initial vertex of the lane boundaries are not visible while it is easy to guess the region where they begin.

\begin{table*}[t]
\centering
\label{table:annot}
\begin{tabular}{c|c|l|c|c||c|l|l|l|}
\cline{2-9} & \multicolumn{4}{c||}{Precision at (cm)} & \multicolumn{4}{c|}{Recall at (cm)} \\ \cline{2-9} 
 & 5   & 10   & 15& 20   & 5  & 10   & 15   & 20   \\ \hline

\multicolumn{1}{|c|}{Ours Before Correction}  &  \textbf{0.195} &  \textbf{0.534} &  \textbf{0.748} & \textbf{0.851}  & 0.169 &  0.461 &  0.647 & 0.736 \\ \hline

\multicolumn{1}{|c|}{Ours After Correction}  & 0.188 &  0.515 &  0.726 &  0.833 &  \textbf{0.189} &  \textbf{0.519} &  \textbf{0.731} &  \textbf{0.841} \\ \hline
\end{tabular}
\caption{Evaluating the annotator in the loop by comparing precision and recall for before and after images with the wrong topology are corrected. On average an annotator takes 1.07 clicks to fix these mistakes.}
\vspace{-4mm}
\end{table*}

\vspace{-2mm}
\paragraph{Qualitative Examples:} 

In Fig. \ref{fig:good}, we demonstrate the abilities of our model to make   high precision and recall predictions with  perfect topology. Our model is able to deal with occlusions due to other vehicles and most importantly has learned to extrapolate the lane boundaries to parts of the image where no evidence exists. Moreover, our model performs well at night where camera based models might have difficulty. We also depict the projection of the predicted lane graph onto the frontal camera view of the road for visualization purposes only.

\vspace{-2mm}
\paragraph{Failure Modes:}

\begin{figure}[t]
\[\arraycolsep=1.0pt
\begin{array}{lll}

\includegraphics[width=0.33\linewidth]{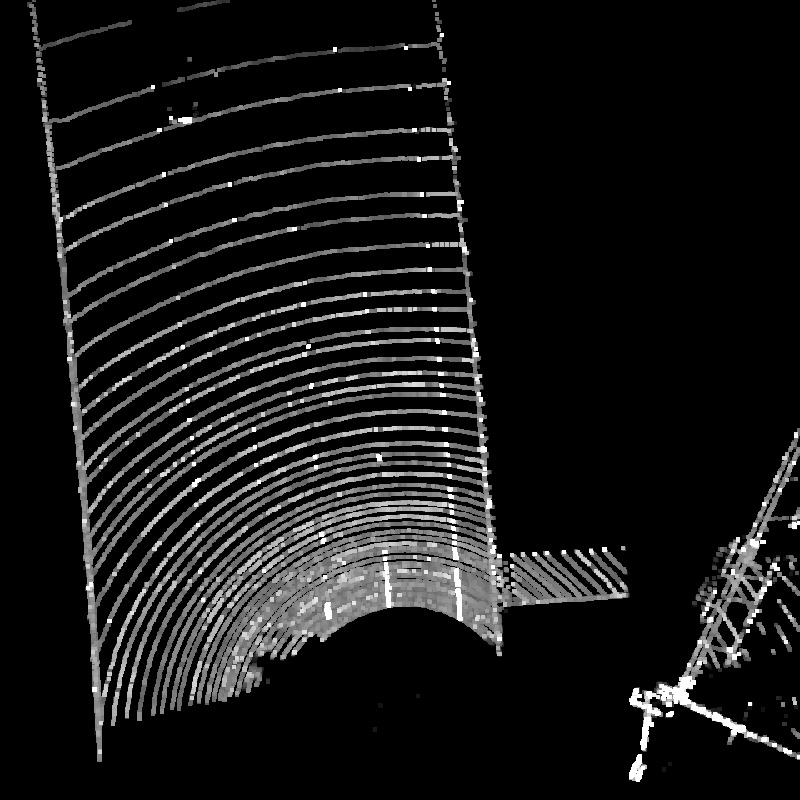}  & 
\includegraphics[width=0.33\linewidth]{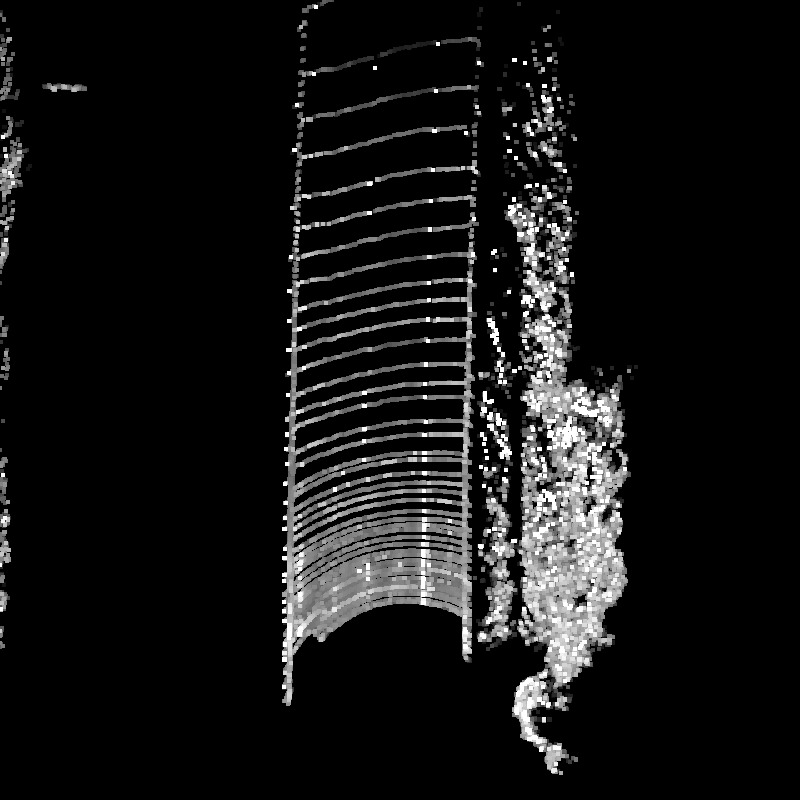}&  
\includegraphics[width=0.33\linewidth]{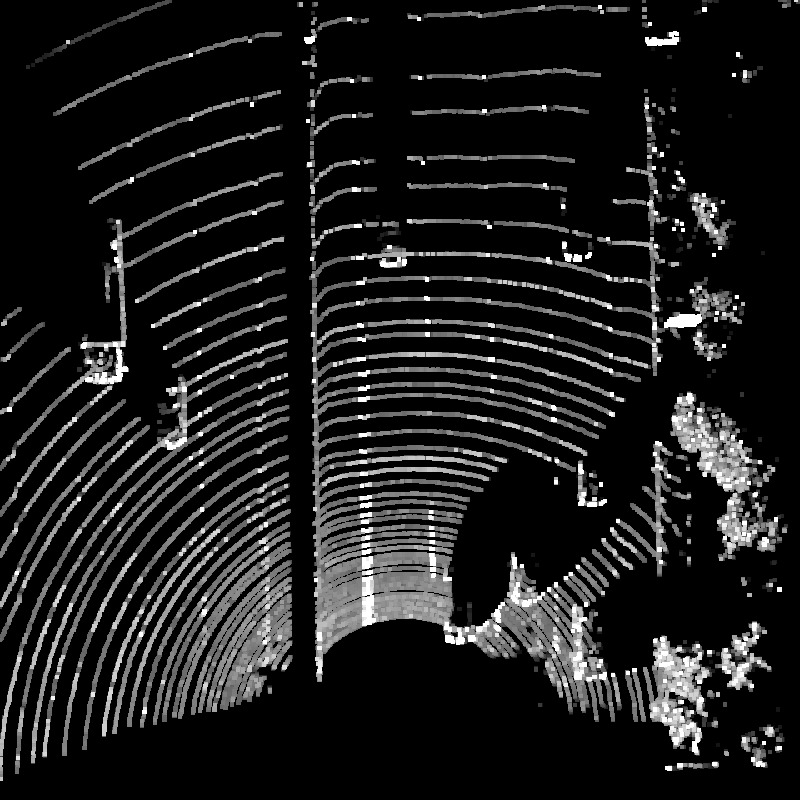} \\

\includegraphics[width=0.33\linewidth]{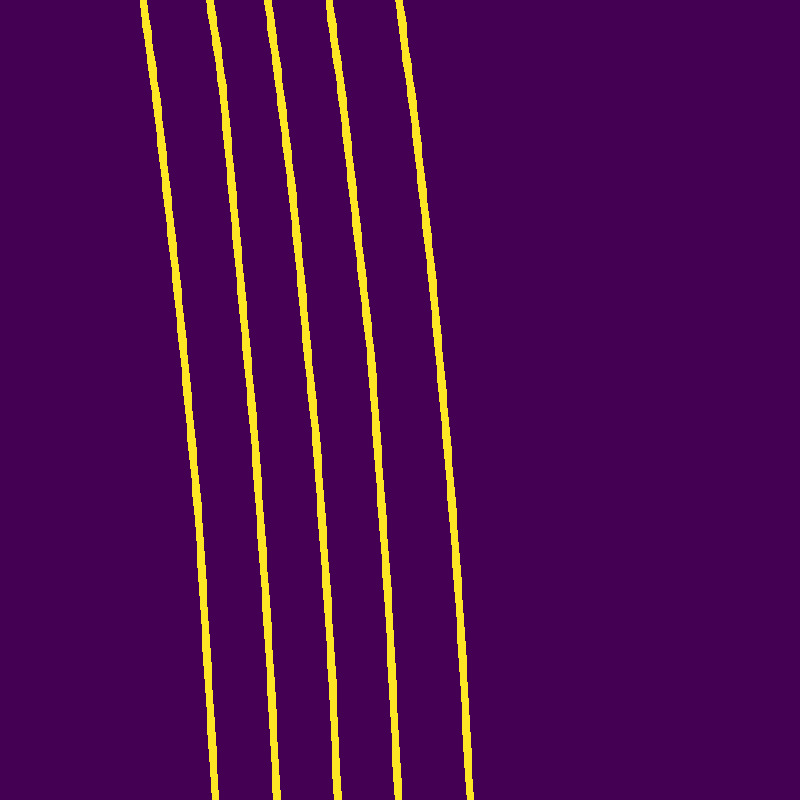}  & 
\includegraphics[width=0.33\linewidth]{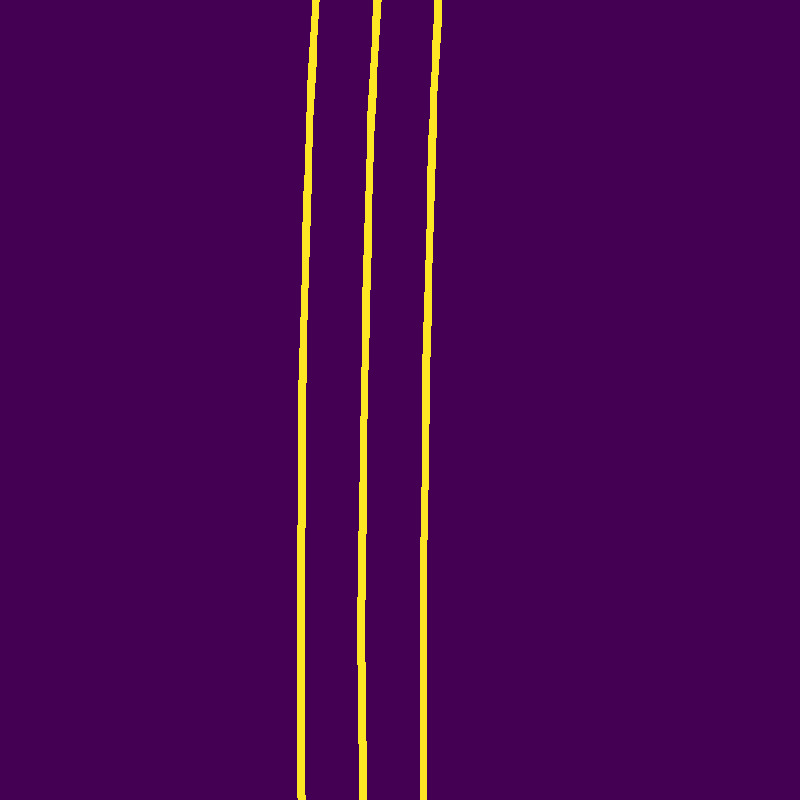}& 
\includegraphics[width=0.33\linewidth]{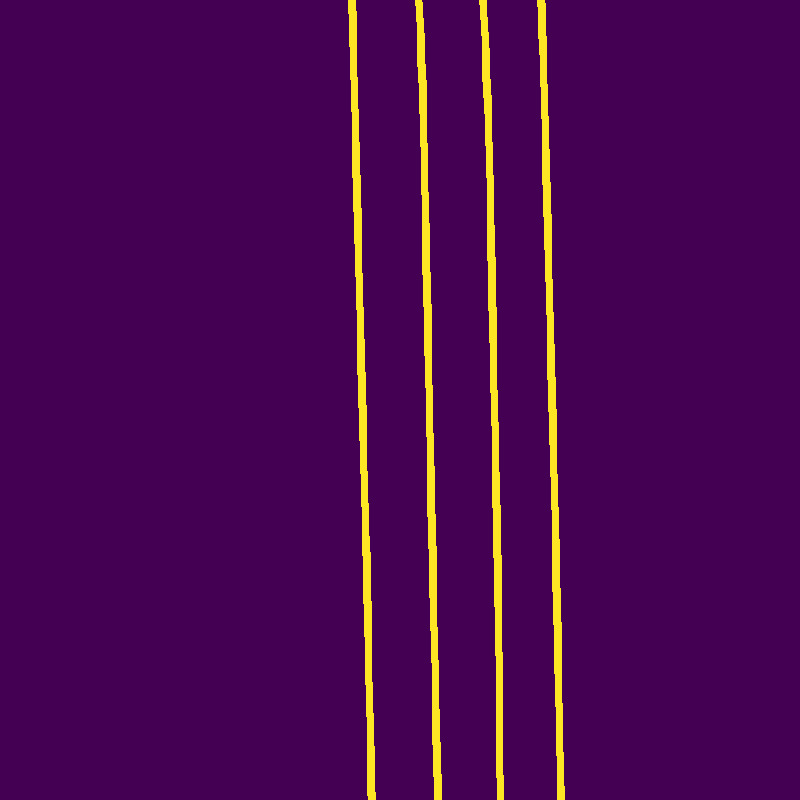} \\

\includegraphics[width=0.33\linewidth]{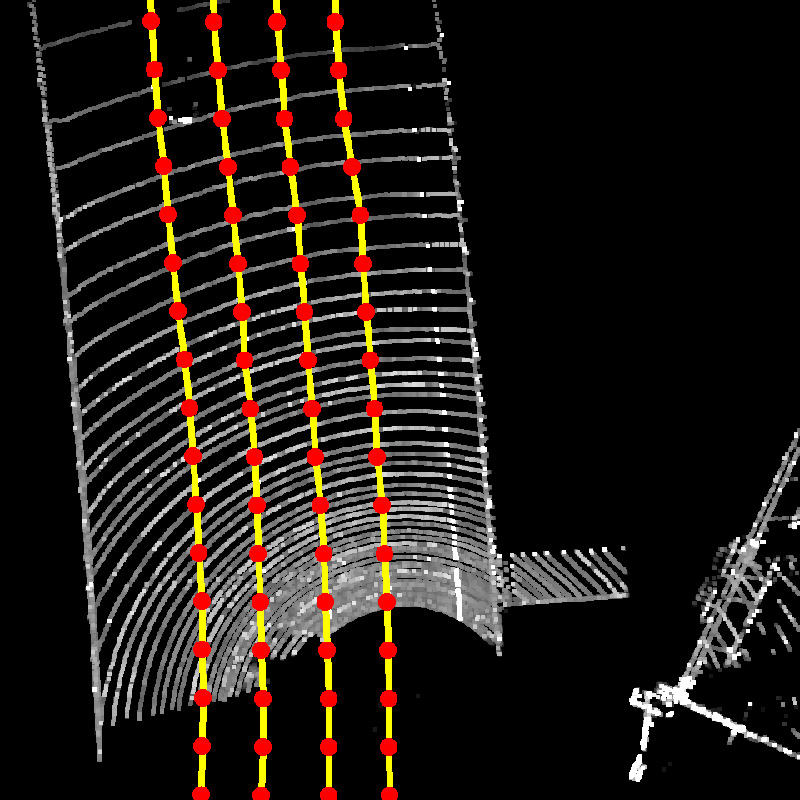}  & 
\includegraphics[width=0.33\linewidth]{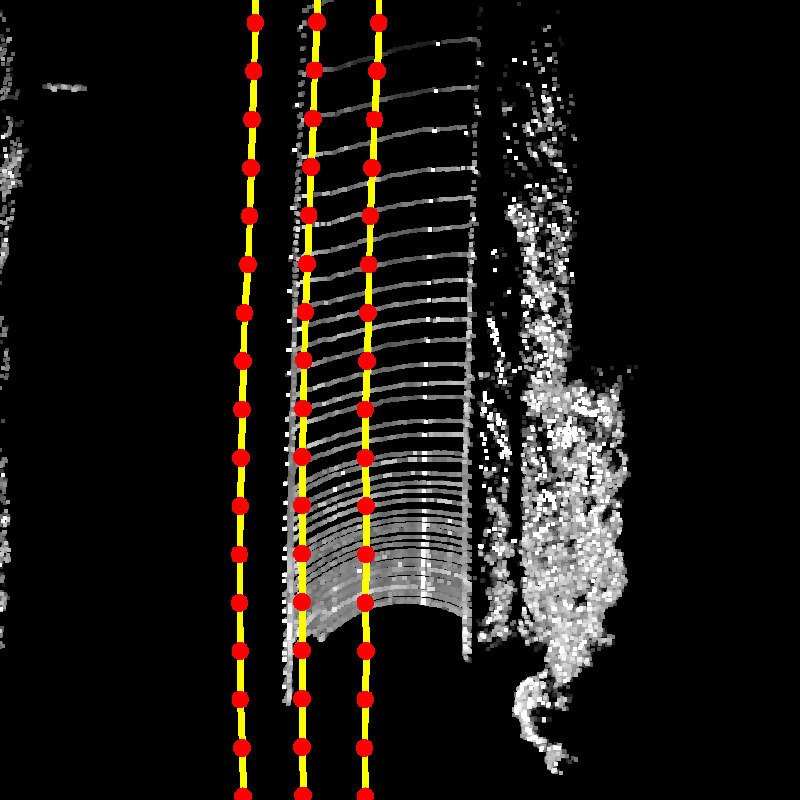}& 
\includegraphics[width=0.33\linewidth]{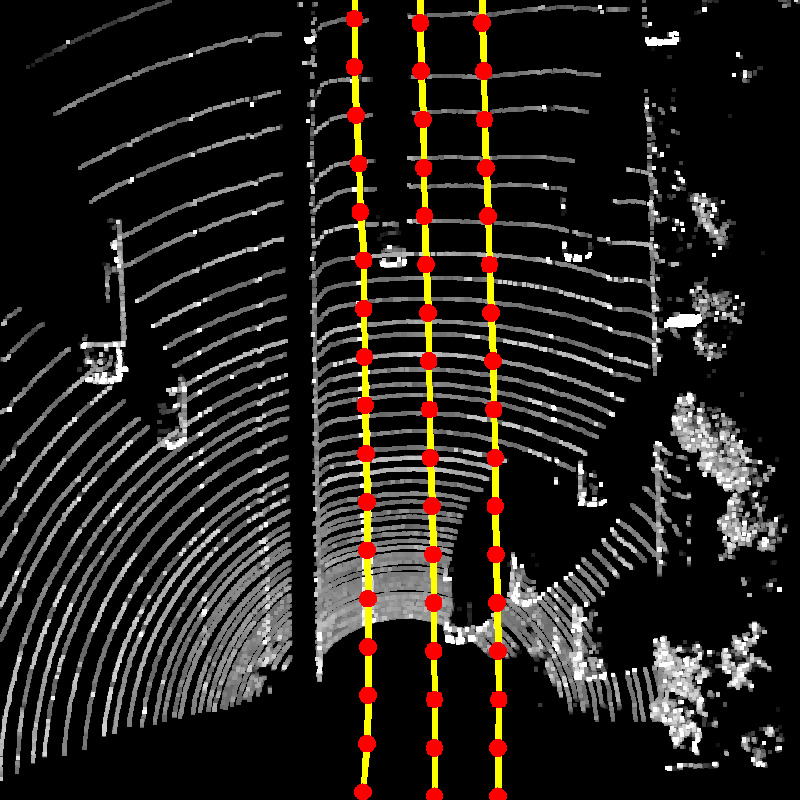} \\

\includegraphics[width=0.33\linewidth, trim={0.8cm 0 0 0}]{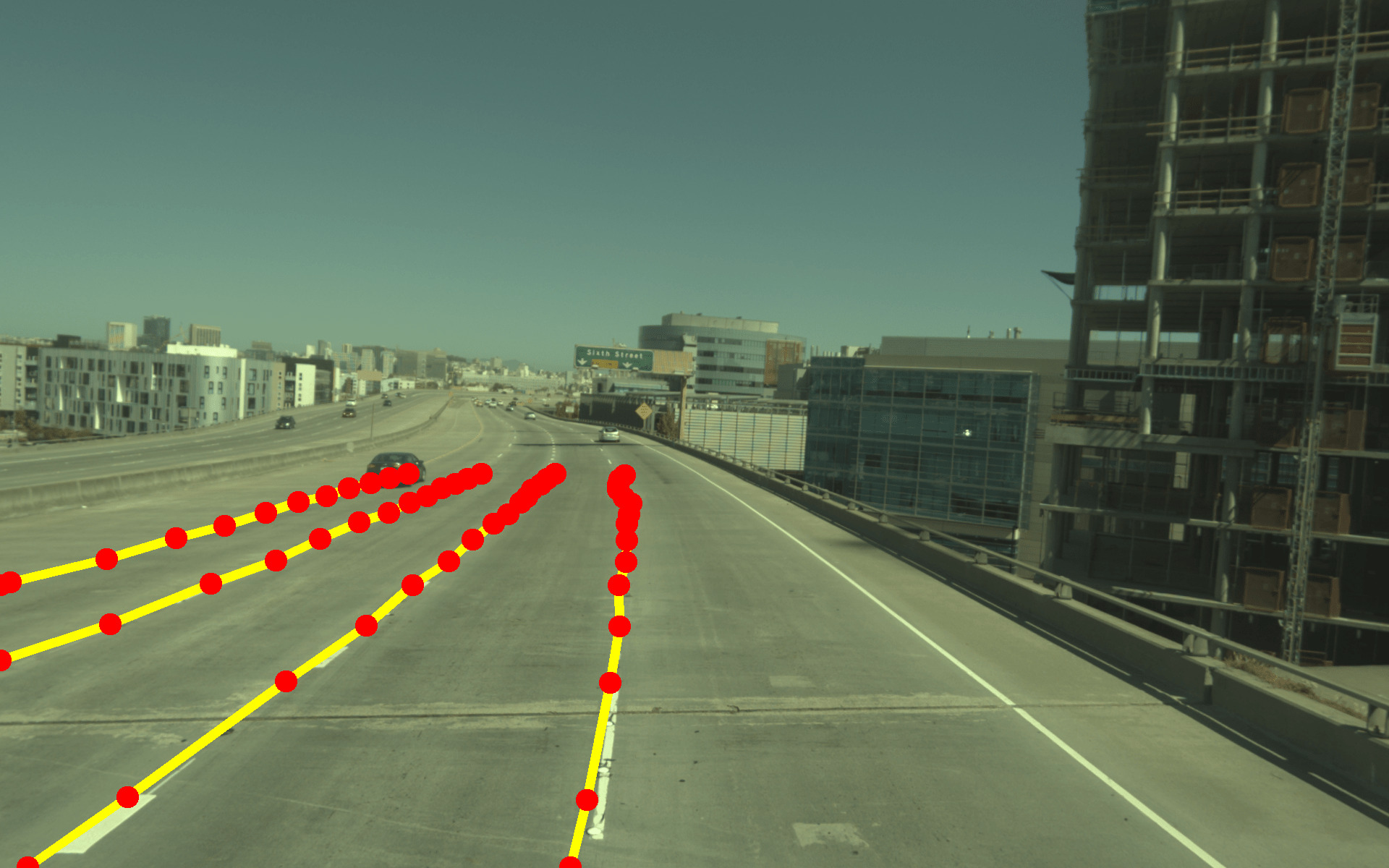}  & 
\includegraphics[width=0.33\linewidth, trim={0.8cm 0 0 0}]{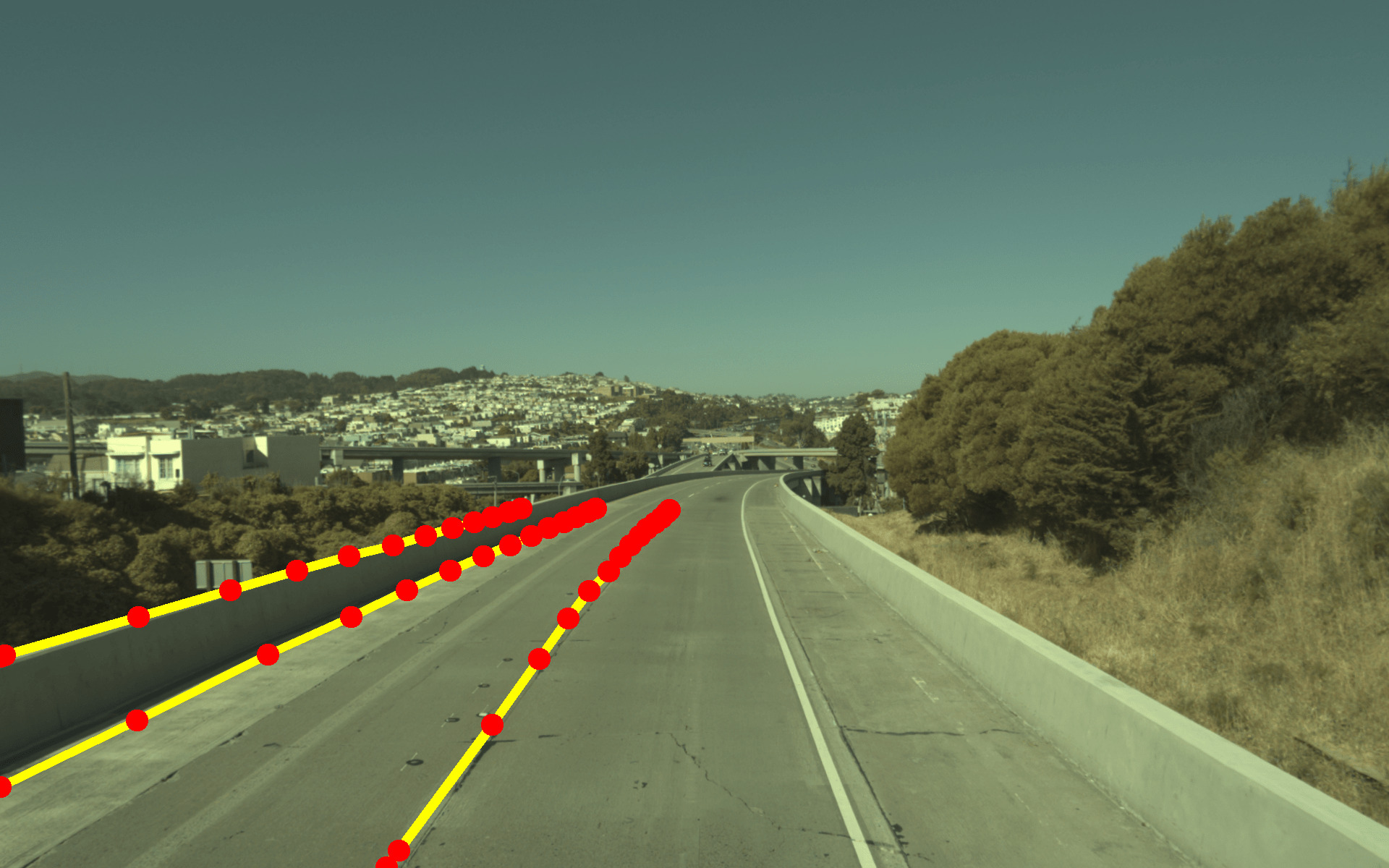}& 
\includegraphics[width=0.33\linewidth, trim={0.8cm 0 0 0}]{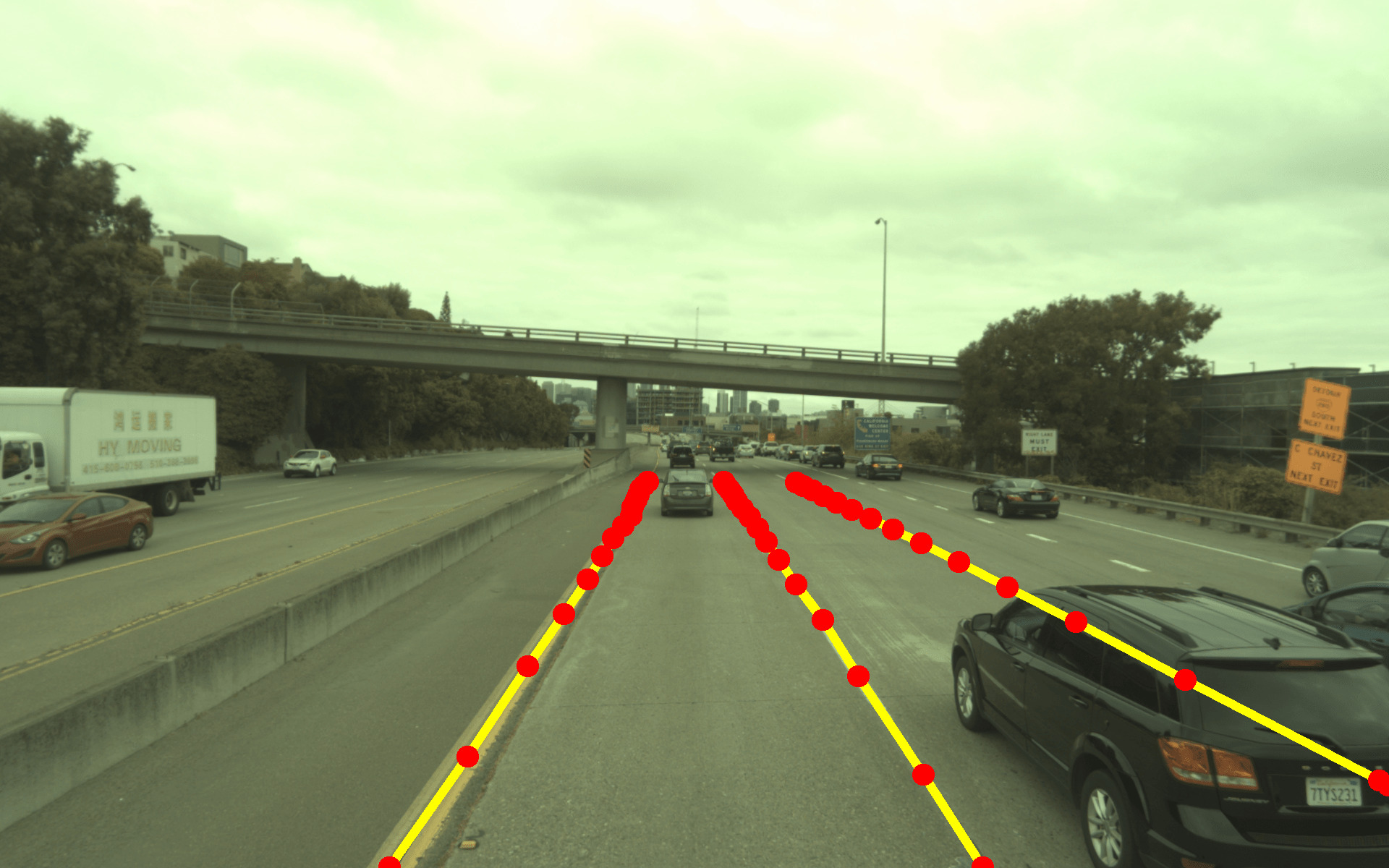}
\end{array}
\]

\caption{Failure Cases. \textbf{Top Row}: Point cloud sweep of the road. \textbf{Second Row}: The ground truth lane graph. \textbf{Third Row}: The lane boundary instances outputs of our network. \textbf{Bottom Row}: The predicted lane boundaries projected onto the image.}
\label{fig:fail}
\vspace{-4mm}
\end{figure}

In Fig. \ref{fig:fail} we visualize a few failure cases. In columns 1 and 3, we observe that the topology is wrong due to an extra predicted lane boundary. In the second column, the road separator is mistaken for a lane boundary. However, note that an annotator can pass through these images and fix the issues with only one click.

\vspace{-2mm}
\paragraph{Inference Time:} 
Our model takes on average 175 ms for one forward pass timed on a Titan XP. 
While the encoder-decoder module takes only 15 ms, the majority of the inference time is spent on the convolution LSTM cells.

\begin{figure}[t]
\[\arraycolsep=1.0pt
\begin{array}{lll}

\includegraphics[width=0.33\linewidth]{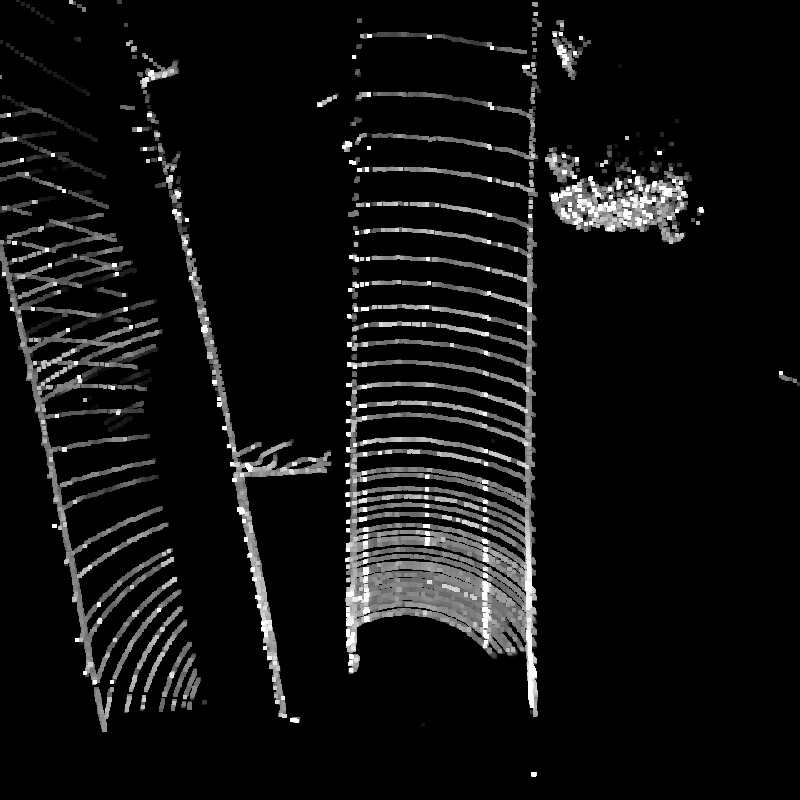}  & 
\includegraphics[width=0.33\linewidth]{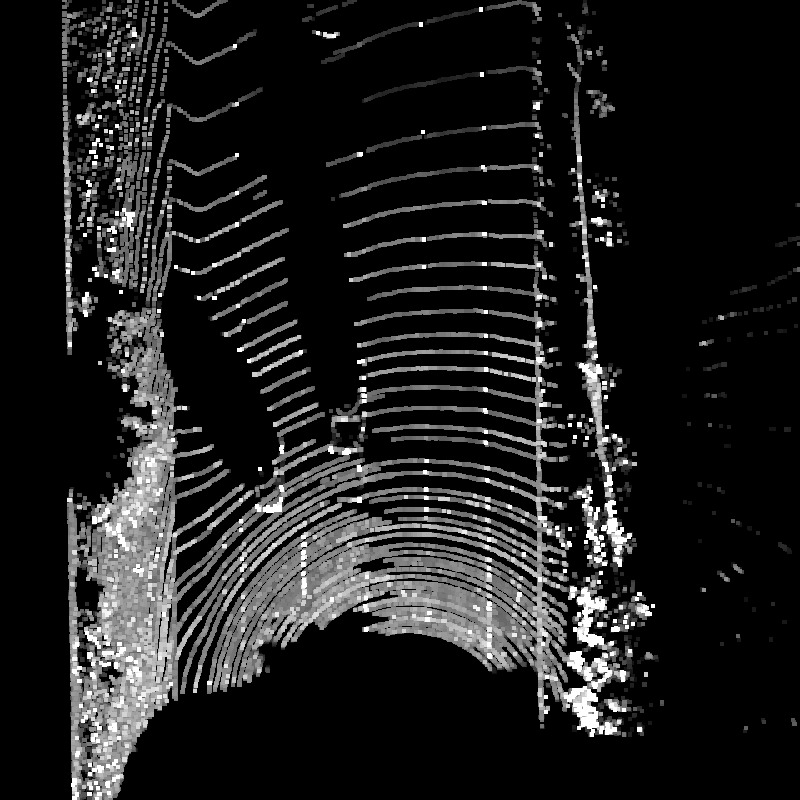}&  
\includegraphics[width=0.33\linewidth]{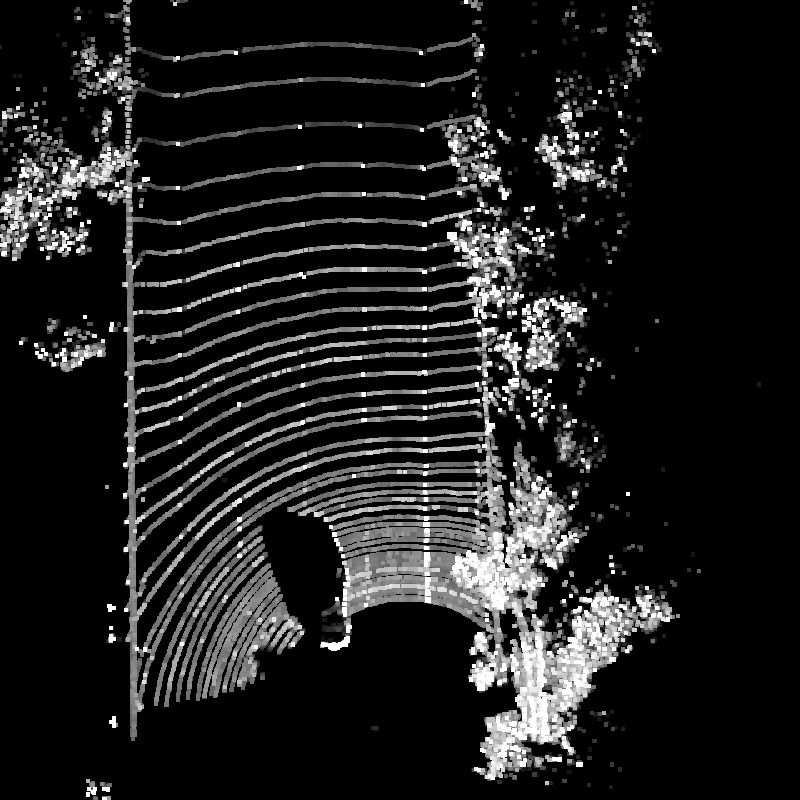} \\

\includegraphics[width=0.33\linewidth]{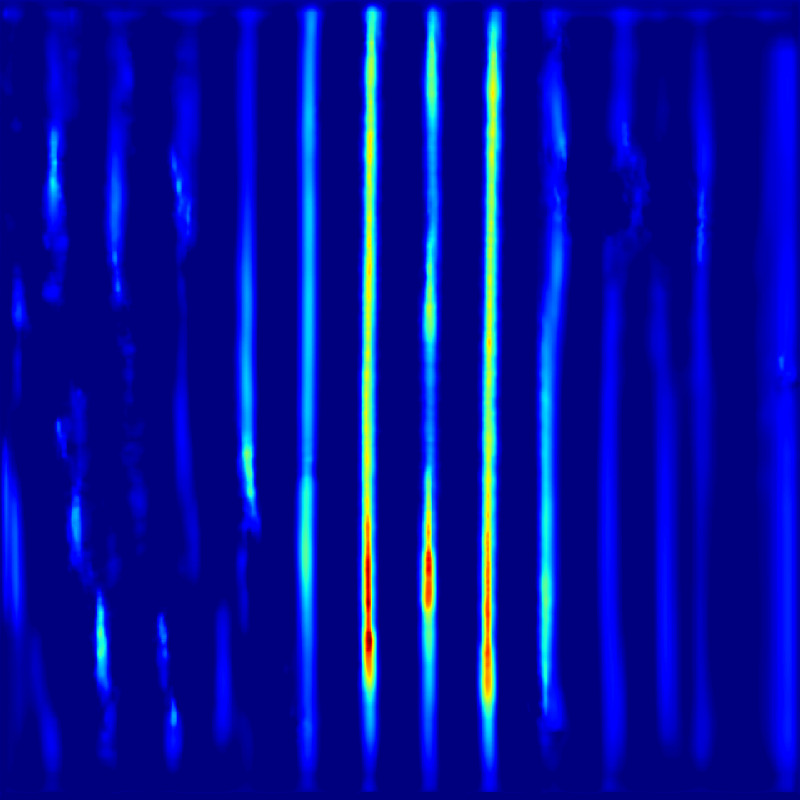}  & 
\includegraphics[width=0.33\linewidth]{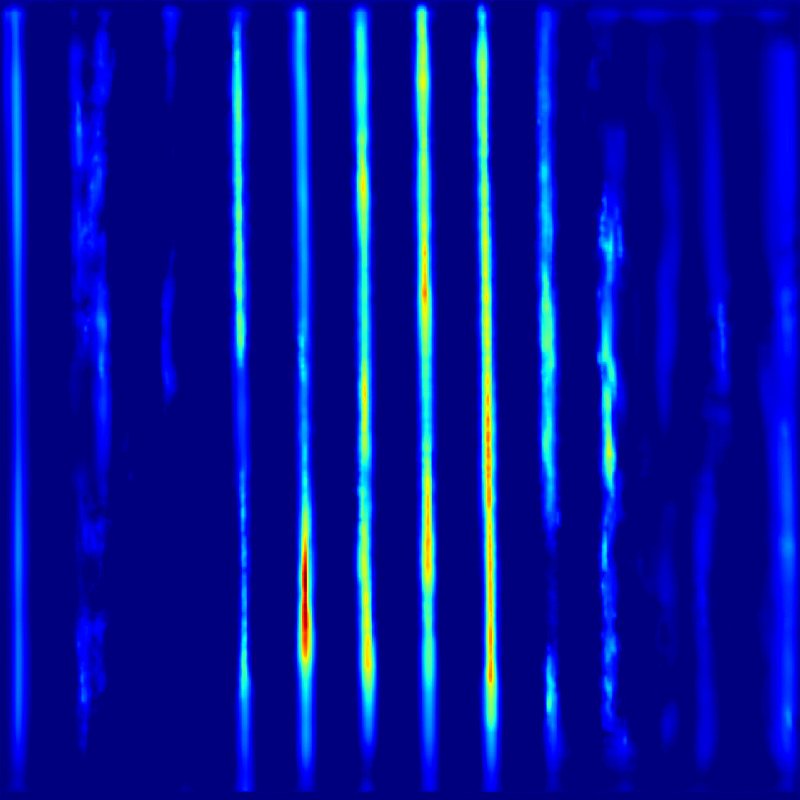}&  
\includegraphics[width=0.33\linewidth]{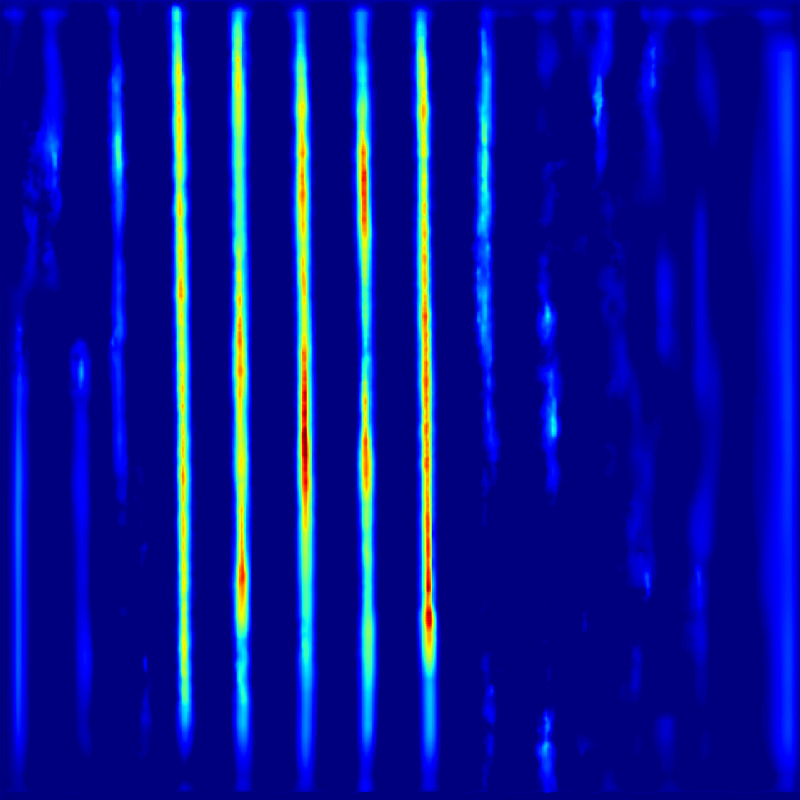} \\

\includegraphics[width=0.33\linewidth]{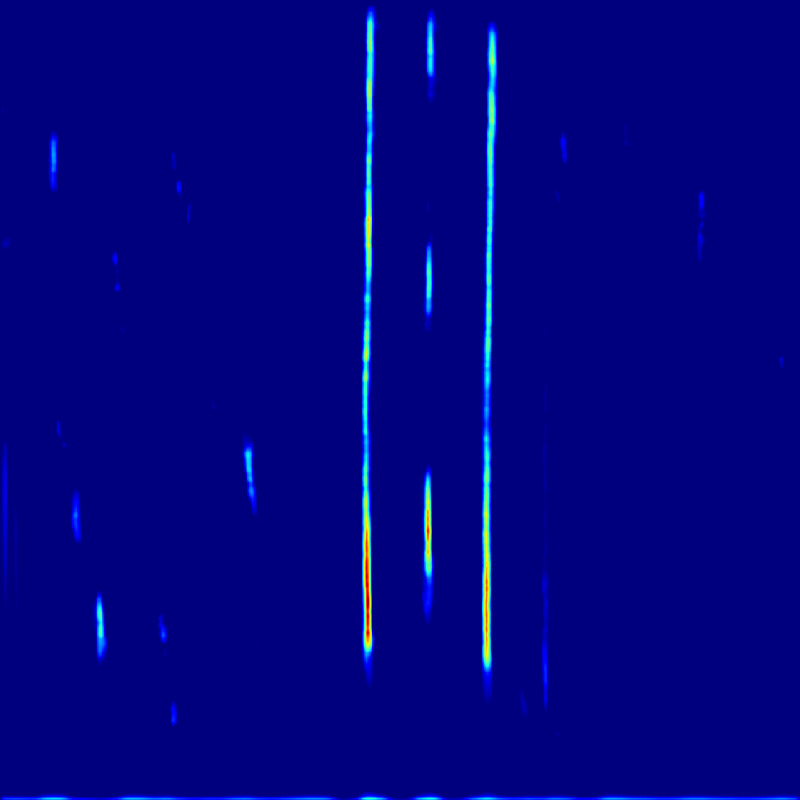}  & 
\includegraphics[width=0.33\linewidth]{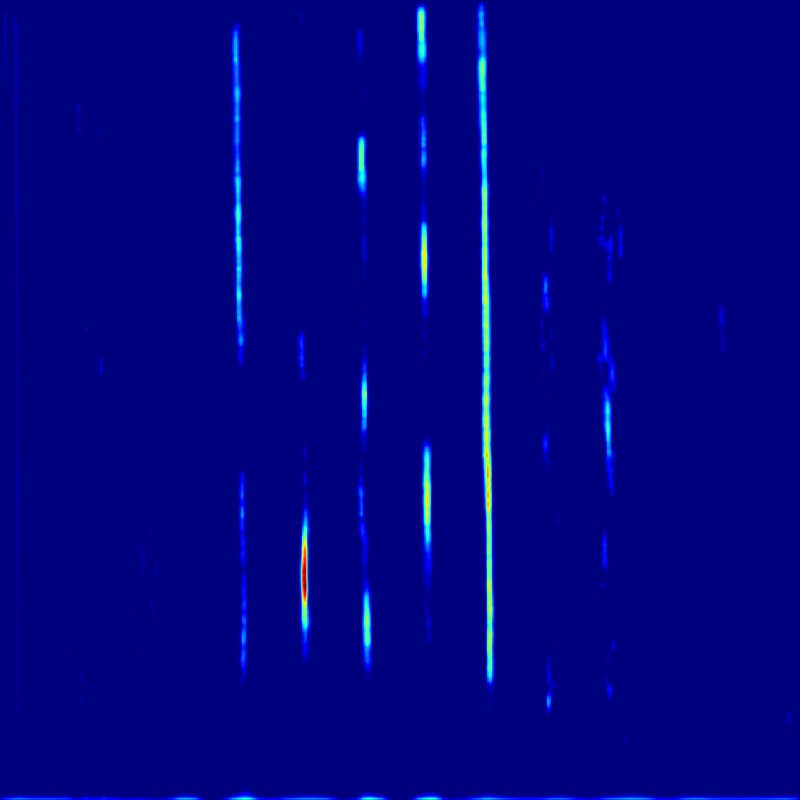}&  
\includegraphics[width=0.33\linewidth]{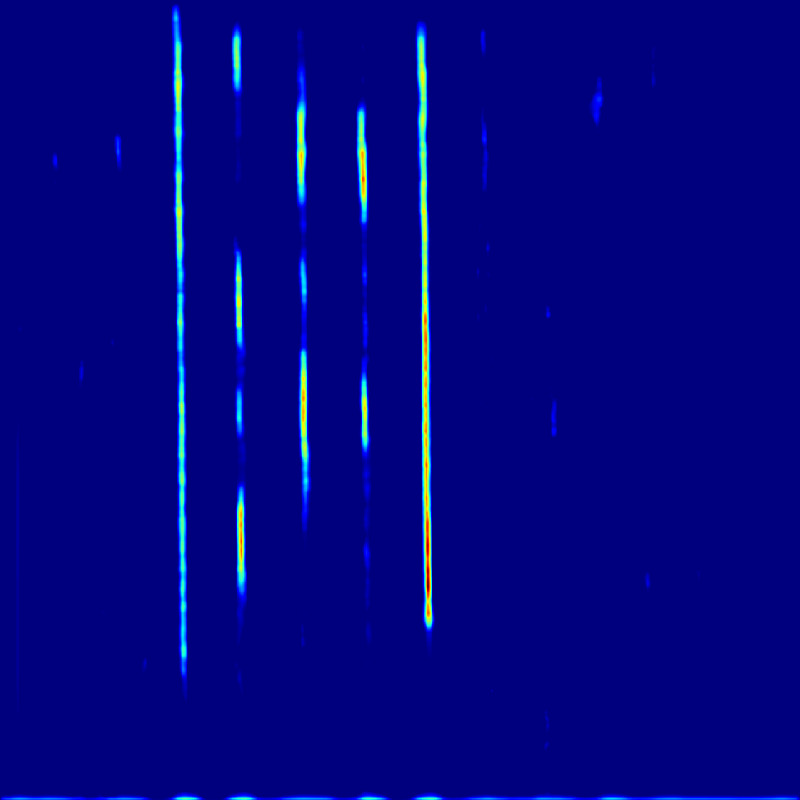} \\

\includegraphics[width=0.33\linewidth, trim={0.8cm 0 0 0}]{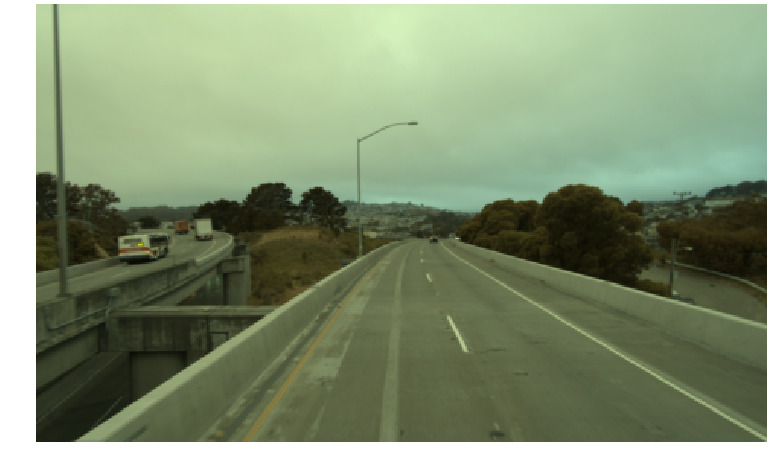}  & 
\includegraphics[width=0.33\linewidth, trim={0.8cm 0 0 0}]{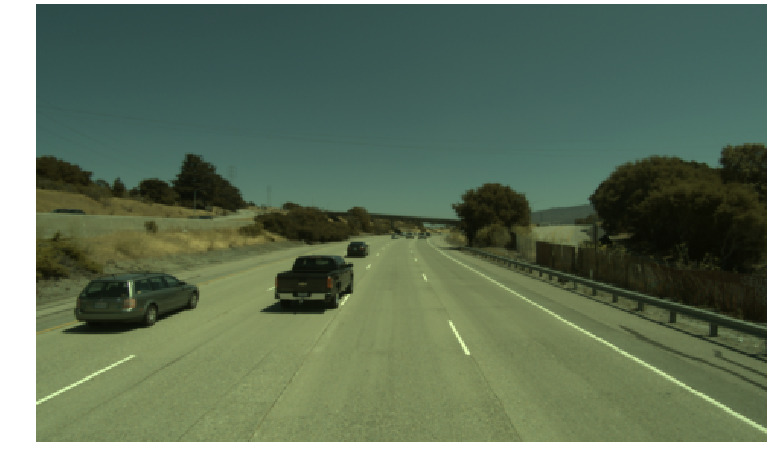}& 
\includegraphics[width=0.33\linewidth, trim={0.8cm 0 0 0}]{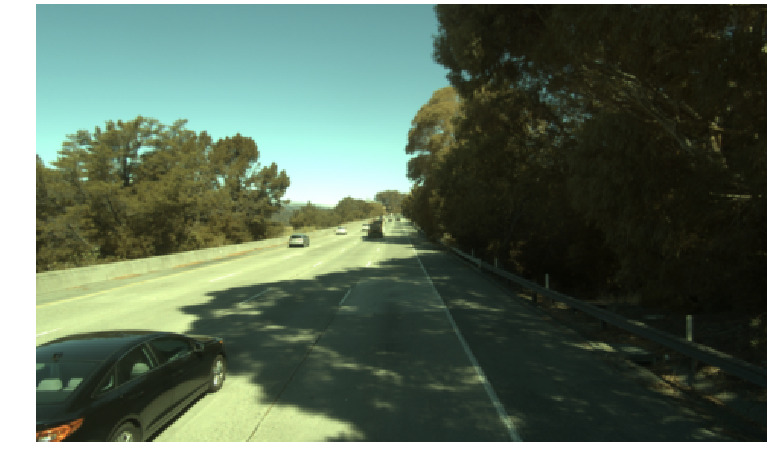}
\end{array}
\]

\caption{Learned Features. \textbf{Top Row}: Point cloud sweep of the road. \textbf{2-4th Rows}: Three feature map channels of the decoder network.  \textbf{Bottom Row}: The camera view of the vehicle.}
\label{fig:feat}
\vspace{-4mm}
\end{figure}

\vspace{-2mm}
\paragraph{Learned Features:} 
In Fig. \ref{fig:feat} we visualize three channels of the last feature map of the decoder network before feeding it is fed into the conv-lstm for lane boundary drawing. The input to the network is a sparse point cloud and the output is a structured lane graph. We observe that in order to learn this mapping, the network has learned to pick out lane markings and   extrapolate them to fine the boundaries. 

%!TEX root = ./top.tex

\section{Conclusion}

In this paper, we proposed a hierarchical recurrent attention network that mimics how an annotator creates a map of the road network. In particular, given a sparse lidar sweep of the road, a recurrent attention module attends to the initial regions of the lane boundaries while a convolutional LSTM draws them out completely. We developed a novel loss function that penalizes the deviation of the edges of the ground truth and predicted polylines rather than their vertices. We demonstrated the effectiveness of our method by extensive experiments on a 90 km of highway.

%-------------------------------------------------------------------------

{\small
\bibliographystyle{ieee}
\bibliography{top}
}

\end{document}